%% file: iclr2027_conference.tex
\documentclass{article} 
\usepackage{iclr2027_conference,times}

\input{math_commands.tex}

\usepackage{hyperref}
\usepackage{url}
\usepackage{tabularx}
\usepackage{booktabs}
\usepackage{multirow}
\usepackage{multicol}
\usepackage{placeins}
\usepackage{dsfont}
\usepackage{makecell}
\usepackage{graphicx}
\usepackage{wrapfig}
\usepackage[table]{xcolor}
\usepackage{array}
\usepackage{enumitem}

\newcolumntype{C}{>{\centering\arraybackslash}X}
\newcolumntype{L}{>{\raggedright\arraybackslash}X}
\newcolumntype{R}{>{\raggedleft\arraybackslash}X}
\newcolumntype{G}{>{\columncolor{gray!15}}C}
\newcolumntype{B}{>{\columncolor{cyan!15}}C}
\newcommand{\ipdcell}[1]{\cellcolor{cyan!15}#1}

\title{Interactive-Policy Distillation with \\Bidirectional Propose-and-Verify}

\author{%
\begin{minipage}{\dimexpr\textwidth-2\tabcolsep\relax}
\vspace{0.35in}
\raggedright\normalfont
\large
\textbf{Shutong Wu}\textsuperscript{1*},
\textbf{Xiwen Chen}\textsuperscript{2},
\textbf{Brendan Rappazzo}\textsuperscript{2}, 
\textbf{Daiheng Zhang}\textsuperscript{3},\\
\textbf{Anderson Schneider}\textsuperscript{2},
\textbf{Yuriy Nevmyvaka}\textsuperscript{2},
\textbf{Jiawei Zhang}\textsuperscript{1}\\[0.6ex]
\normalsize
\textsuperscript{1}Department of Computer Sciences, University of Wisconsin--Madison\\
\textsuperscript{2}Machine Learning Research, Morgan Stanley\\
\textsuperscript{3}Department of Electrical and Computer Engineering, Rutgers University--New Brunswick
\end{minipage}%
}

\newcommand{\authorcontactnotes}{%
  \begingroup
  \renewcommand{\thefootnote}{\fnsymbol{footnote}}%
  \footnotetext[1]{Work done during the internship at Morgan Stanley. Correspondence to Shutong Wu \textless{}\nolinkurl{swu494@wisc.edu}\textgreater{} and Jiawei Zhang \textless{}\nolinkurl{jzhang2924@wisc.edu}\textgreater{}. Code available at \url{https://github.com/cychomatica/Interactive-Policy-Distillation}.}%
  \endgroup
}

\iclrpreprintcopy 
\begin{document}

\maketitle
\ificlrfinal
  \authorcontactnotes
\else\ificlrpreprint
  \authorcontactnotes
\fi\fi

\begin{abstract}
On-policy distillation (OPD) trains a student model on its self-generated trajectories with dense token-level teacher feedback. However, naive OPD may suffer from teacher unanchoring, where the student's reasoning trajectory drifts far from the teacher, causing the teacher to be queried on states it would hardly visit and thus provide unreliable supervision.
We propose \textbf{Interactive-Policy Distillation (IPD)}, which applies adaptive teacher intervention to the student rollout. Under a bidirectional propose-and-verify state machine, the student and teacher alternately exchange their roles as proposer and verifier, and collaboratively generate mixed-source trajectories.
Then different supervisions are applied according to the source of each token. This bidirectional propose-and-verify mechanism and the source-split loss make IPD not only a more performant distillation method, but also a unified bridge between on-policy and off-policy paradigms. 
To make the interleaved dual-model rollouts more efficient, we also design a dedicated fused inference engine that co-hosts both models in one serving instance with separate KV caches and instantiates the state machine model to distribute, collect, and process requests. 
On math reasoning tasks and across multiple teacher-student model pairs, student models trained with IPD not only outperform those trained with OPD, but also demonstrate higher data efficiency. Specifically, when distilling Qwen3-30B-A3B into Qwen3-1.7B-Base, IPD brings a $+3.28$ mean@8 and a $+3.28$ best@8 benchmark-averaged accuracy improvement compared with OPD. Besides, IPD only consumes about $1/4$ of the training examples and steps to outperform OPD trained on the whole training dataset for one epoch. 
We also investigate the impact of different loss variants and takeover / handback configurations, and demonstrate the robustness of IPD on different training data. 
\end{abstract}
\section{Introduction} 
\label{sec:intro}

Large language models (LLMs) have advanced at a remarkable pace over the past few years, and current frontier models attain expert-level performance in mathematics, programming, and scientific problems, and drive increasingly capable agents \citep{openai2026astra, anthropic2026fable, zai2026glm53, xu2026deepseek}. These extraordinary capabilities, however, are usually obtained by the very largest models, which pair over hundreds of billions of parameters and long chains of thought. The high memory and computation cost makes them a suboptimal fit for latency-sensitive, on-device, or high-throughput deployments.
Knowledge distillation~\citep{hinton2015distilling} offers a way out by transferring as much of the large model's capability as possible into a more compact and cheaper student model. In the LLM era, the dominant paradigm used to be off-policy distillation via supervised fine-tuning (SFT), where the teacher generates high-quality reasoning traces, and the student learns to imitate them token by token \citep{kim2016sequence, ho2023large, magister2023teaching, hsieh2023distilling}. The distilled variants of DeepSeek-R1, trained by SFT on roughly 800K teacher-generated samples, are a prominent recent example of how far this simple recipe scales~\citep{guo2025deepseek}.

Despite its scalability, off-policy distillation faces a structural challenge: the student is trained under the teacher's demonstrations, and the imitation is only ever supervised on teacher-anchored states that the student is not likely to visit. Therefore, the student never learns to continue from its own partial generations, and small early deviations can compound at inference time, as discussed in the classic exposure-bias problem \citep{bengio2015scheduled, ranzato2015sequence, arora2022exposure}. 
Long and stylized teacher traces can exceed what a small student is able to absorb \citep{mirzadeh2020improved, shing2025taid, li2025small}, and the forward KL pushes the student to spread probability mass over every teacher mode rather than reliably on modes it can actually realize \citep{gu2024minillm}. 
On-policy distillation (OPD) overcomes these challenges by moving the training distribution to the student. The student generates its own responses, and the teacher supervises every generated token with dense probabilistic feedback \citep{gkd, gu2024minillm, lu2025onpolicydistillation}. Training on self-generated states removes the training-inference mismatch by construction. The student learns from states it actually visits \citep{ross2011reduction}, while the teacher's per-token distributions provide denser learning signals than scalar outcome rewards of reinforcement learning with verifiable rewards (RLVR). With those advantages, OPD can match RLVR-based training with much less compute \citep{lu2025onpolicydistillation}, and has quickly moved into production practice as a core stage of post-training pipelines for building reasoning models \citep{yang2025qwen3, ma2026mopd}, accompanied by a growing line of refinements on divergence choice and data mixing \citep{ko2024distillm}.

\begin{figure}[!h]
  \includegraphics[width=0.99\textwidth]{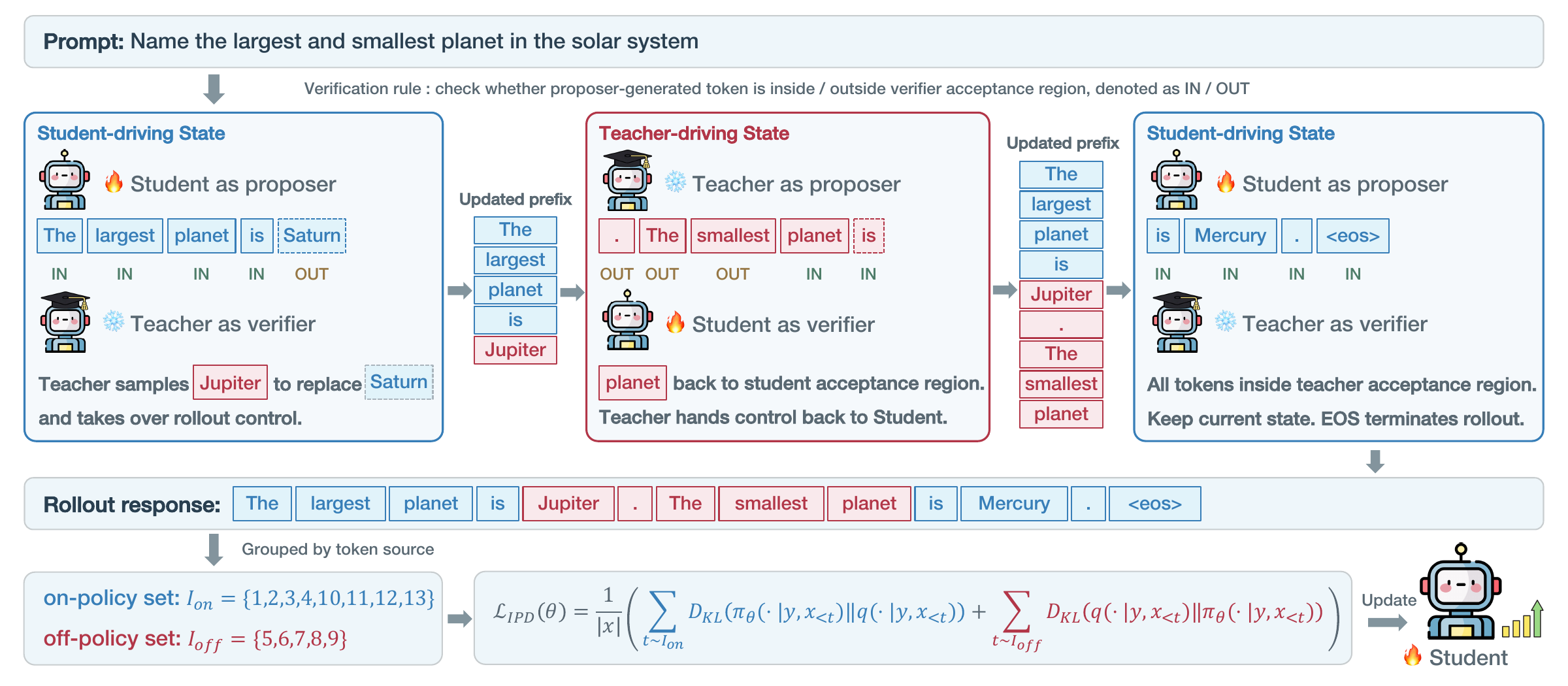}
  \caption{The rollout and training demonstration of IPD. The student and teacher collaboratively generate the rollout under a bidirectional propose-and-verify state machine. Then tokens are grouped by their sources, and the IPD source-split loss is applied to update the student weights.}
  \label{fig:ipd_demo}
\end{figure}

However, student-generated trajectories can deviate far from the teacher when there is a large reasoning gap between them, and then the teacher is queried by states it would hardly visit and may provide unreliable learning signals \citep{ko2024distillm, skd}. We call this failure mode \emph{teacher unanchoring}. 
To overcome this challenge, we introduce \textbf{Interactive-Policy Distillation (IPD)}, which couples a bidirectional propose-and-verify state machine with different supervisions on tokens from different sources. As shown in Figure~\ref{fig:ipd_demo}, IPD allows the teacher to dynamically correct a drifting rollout and returns control once the corrected continuation falls within the student's acceptance region where student can actually resume. 
As the resulting trajectory interleaves the student's own decisions with teacher-provided corrective demonstrations, IPD assigns different supervisions by token sources, \emph{i.e.}, reverse KL at student-generated positions and forward KL at teacher-generated positions. Our contributions are summarized as follows:
\begin{itemize}[leftmargin=*]
    \setlength{\itemsep}{1pt}
    \setlength{\parsep}{0pt}
    \item We introduce IPD, a novel distillation algorithm with adaptive teacher intervention through a bidirectional propose-and-verify state machine, in which student and teacher dynamically exchange proposer and verifier roles. Its source-split objective applies reverse KL to student-generated positions and forward KL to teacher-generated positions, bridging on-policy and off-policy distillation paradigms.
    \item We design a fused inference engine, which co-hosts both models in one serving instance, maintains separate KV caches, and schedules requests according to their current proposer and verifier roles, to achieve more efficient dual-model scheduling during the rollout stage. 
    \item Empirically, across four teacher-student pairs on mathematical reasoning, IPD achieves the highest benchmark-averaged mean@8 and best@8 accuracy and leads on most individual benchmark metrics. Besides, IPD shows higher data efficiency. When distilling Qwen3-30B-A3B to Qwen3-1.7B-Base, IPD surpasses the mean@8 of OPD trained for a full epoch (279 steps) after only 60 training steps. We also demonstrate the robustness of IPD across different training datasets.
\end{itemize}

\section{Related Works}
\label{sec:related_works}

\noindent\textbf{On-policy distillation for LLMs.} LLM distillation used to be dominated by off-policy distillation on teacher-generated traces \citep{hinton2015distilling, kim2016sequence, guo2025deepseek}. The main weakness of off-policy distillation is that the student trains on teacher-authored prefixes but deploys on its own, and is vulnerable to exposure bias \citep{bengio2015scheduled, arora2022exposure}. 
GKD \citep{gkd} applies purely on-policy student sequences for training, and MiniLLM \citep{gu2024minillm} studies sequence-level reverse KL on those sequences. DistiLLM \citep{ko2024distillm} introduces skew divergences with an adaptive off-policy scheduler, and DistiLLM-2 \citep{ko2025distillm} contrasts teacher and student responses with skew forward and reverse KL.
Recently, OPD has already been incorporated as a core stage into reasoning-model post-training pipelines \citep{yang2025qwen3, lu2025onpolicydistillation, xiao2026mimo}.

\noindent\textbf{Challenges in on-policy distillation.}
Recent analyses characterize unreliable teacher feedback on student rollouts \citep{fu2026revisiting, armandpour2026unmasking, li2026rethinking}. EOPD \citep{jin2026entropyaware} adjusts the loss while retaining student-generated rollouts. TrOPD \citep{xing2026trust} uses a teacher-generated prefix followed by a student continuation and applies region-specific losses, but without dynamic teacher intervention during the continuation. SKD \citep{skd} adapts speculative propose-and-verify \citep{leviathan2023fast, chen2023accelerating} to rollout and implements dynamic teacher corrections. The student and teacher remain fixed roles as proposer and verifier. When a student token is rejected, the teacher resamples that position, and the student resumes generation from the corrected prefix.
Interactive imitation learning studies intervention and return of control through human judgment in HG-DAgger \citep{kelly2019hg} or uncertainty-based gating in EnsembleDAgger \citep{menda2019ensembledagger}. 
In contrast, IPD build a state machine model that dynamically swaps proposer and verifier roles through teacher takeover on rejected student tokens and handback on accepted teacher tokens, and then trains the resulting mixed-source trajectory with a source-split objective.

\section{Methodology}
\label{sec:method}

\subsection{Preliminaries}

Let $y\sim\mathcal{D}$ be a prompt from a dataset $\mathcal{D}$, $x=(x_1,\ldots,x_T)$ a response trajectory, and $c_t=(y,x_{<t})$ the prefix context at step $t$. The student and teacher next-token distributions are denoted as $\pi_\theta(\cdot\mid c_t)$ and $q(\cdot\mid c_t)$, and their complete-response probabilities factorize as $\pi_\theta(x\mid y)=\prod_{t=1}^{|x|}\pi_\theta(x_t\mid c_t)$ and $q(x\mid y)=\prod_{t=1}^{|x|}q(x_t\mid c_t)$.


\noindent\textbf{Off-Policy Distillation.}
Traditional knowledge distillation trains on prefixes sampled from a distribution $\mu$, which is typically a dataset of fixed reference responses or trajectories sampled from a teacher model \citep{hinton2015distilling, kim2016sequence}.  
Its standard objective is minimizing the forward KL, 
\begin{equation}
    \mathcal{L}_{\mathrm{off}}(\theta)
    = \mathbb{E}_{\substack{y\sim\mathcal{D}, x\sim \mu(\cdot\mid y)}}
    \left[
    \frac{1}{|x|}\sum_{t=1}^{|x|}
    D_{\mathrm{KL}}\!\left(
        q(\cdot\mid c_t)\,\Vert\,\pi_\theta(\cdot\mid c_t)
    \right)
    \right].
    \label{eq:off_policy_kd}
\end{equation}
The fixed prefixes make optimization stable and permit dense supervision over the vocabulary, but they also induce a state-distribution mismatch: at inference time, the student must condition on its own previous tokens, including mistakes that are absent from teacher-generated trajectories.

\noindent\textbf{On-Policy Distillation.}
Compared with off-policy distillation, on-policy distillation (OPD) evaluates teacher feedback on responses sampled from the student itself \citep{gkd, gu2024minillm}. Under the notations above, its theoretical objective is the sequence-level reverse KL
\begin{equation}
    \mathcal{L}^{\mathrm{seq}}_{\mathrm{on}}(\theta)
    = \mathbb{E}_{y\sim\mathcal{D}}
    \left[
      D_{\mathrm{KL}}\!\left(
        \pi_\theta(\cdot\mid y)\,\Vert\,q(\cdot\mid y)
      \right)
    \right] 
    = \mathbb{E}_{\substack{y\sim\mathcal{D}, x\sim\pi_\theta(\cdot\mid y)}}
    \left[
      \sum_{t=1}^{|x|} D_{\mathrm{KL}}\!\left(
        \pi_\theta(\cdot\mid c_t)\,\Vert\,q(\cdot\mid c_t)
      \right)
    \right],
    \label{eq:on_policy_kd}
\end{equation}
The outer expectation samples prefixes from the student-induced state distribution, while each summand is a token-level conditional reverse KL at one such prefix. 
At a fixed prefix $c_t$, the exact token-level conditional reverse KL is computed over the entire vocabulary $\mathcal{V}$ as
\begin{equation}
    \ell^{\mathrm{full}}_t(\theta) 
    = D_{\mathrm{KL}}\!\left(
      \pi_\theta(\cdot\mid c_t)\,\Vert\,q(\cdot\mid c_t)
      \right) 
    = \sum_{v\in\mathcal{V}}\pi_\theta(v\mid c_t) \log\frac{\pi_\theta(v\mid c_t)}{q(v\mid c_t)}.
    \label{eq:full_vocab_opd}
\end{equation}
Optimizing Eq.~(\ref{eq:full_vocab_opd}) requires forward and backward passes over all $|\mathcal{V}|$ indices at every position, which is computationally expensive \citep{li2026rethinking, oh2026klforkl}.

\noindent\textbf{Sampled-Token and Top-$K$ OPD.}
As a less expensive alternative to the full-vocabulary KL in Eq.~(\ref{eq:full_vocab_opd}), sampled-token OPD evaluates only the student-sampled token $x_t\sim\pi_\theta(\cdot\mid c_t)$. The resulting single-sample Monte Carlo estimate of the token-level KL value at the fixed prefix $c_t$ is
\begin{equation}
    \widehat{\ell}^{\mathrm{sample}}_t
    =\log \pi_\theta(x_t\mid c_t)-\log q(x_t\mid c_t).
    \label{eq:sampled_token_opd}
\end{equation}

Although Eq.~(\ref{eq:sampled_token_opd}) is an unbiased estimate for the full-vocabulary KL, the variance introduced by sampling a single token can make updates noisy and unstable \citep{oh2026klforkl}. 
Top-$K$ OPD reaches a sweet spot between full-vocabulary and sampled-token OPD. It retains multi-token supervision on a small high-probability support while avoiding a full-vocabulary computation \citep{li2026rethinking, fu2026revisiting}. 
Let
\begin{equation}
    S_t^{\pi}=\operatorname{Top}_K\!\left(
      \pi_\theta(\cdot\mid c_t)\right),
    \qquad
    S_t^{q}=\operatorname{Top}_K\!\left(q(\cdot\mid c_t)\right),
    \label{eq:topk_sets}
\end{equation}
denote the student and teacher top-$K$ supports, respectively, where $\operatorname{Top}_K$ returns the $K$ highest-probability indices, and denote $S_t\subseteq\mathcal{V}$ as the support for distillation, which can usually be $S_t^{\pi}$, $S_t^{q}$, $S_t^{\pi} \cup S_t^{q}$, or $S_t^{\pi} \cap S_t^{q}$. As the two models generally assign different total probability mass to $S_t$, each distribution should be normalized by 
\begin{equation}
    \bar{\pi}^{S_t}_\theta(v\mid c_t) = \frac{\pi_\theta(v\mid c_t)\mathds{1}[v\in S_t]}{\sum_{u\in S_t}\pi_\theta(u\mid c_t)}, \quad
    \bar{q}^{S_t}(v\mid c_t) = \frac{q(v\mid c_t)\mathds{1}[v\in S_t]}{\sum_{u\in S_t}q(u\mid c_t)}.
    \label{eq:topk_normalization}
\end{equation}
Then, the token-level top-$K$ KL and its rollout-level aggregated loss are given by
\begin{gather}
    \ell^{\mathrm{top}\text{-}K}_t(\theta;S_t)
    =D_{\mathrm{KL}}\!\left(
      \bar{\pi}^{S_t}_\theta(\cdot\mid c_t)
      \,\Vert\,\bar{q}^{S_t}(\cdot\mid c_t)\right) = \sum_{v\in S_t}\bar{\pi}^{S_t}_\theta(v\mid c_t) \log\frac{\bar{\pi}^{S_t}_\theta(v\mid c_t)} {\bar{q}^{S_t}(v\mid c_t)}, \label{eq:topk_opd_on_sequence}\\
    \mathcal{L}^{\mathrm{top}\text{-}K}_{\mathrm{on}}(\theta)
    =\mathbb{E}_{\substack{y\sim\mathcal{D}, x\sim\pi_\theta(\cdot\mid y)}}
      \left[\sum_{t=1}^{|x|} \ell^{\mathrm{top}\text{-}K}_t(\theta;S_t)\right].
    \label{eq:topk_opd}
\end{gather}
Although top-$K$ OPD is a biased approximation to full-vocabulary OPD as it discards probability mass outside $S_t$, it replaces the high-variance single-token estimate of sampled-token OPD with deterministic multi-token supervision over the selected high-probability support for each fixed prefix $c_t$, and therefore provides a practical middle ground with denser and more stable supervision than sampled-token OPD \citep{fu2026revisiting}, while at the same time reducing the per-token computation and backward-pass cost of full-vocabulary OPD \citep{li2026rethinking, oh2026klforkl}.

\subsection{Interactive-Policy Distillation}

\begin{figure}[!t]
  \centering
  \includegraphics[width=0.6\textwidth]{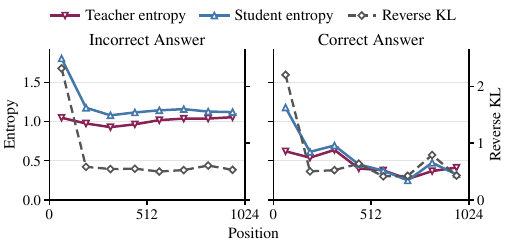}
  \caption{Average entropy and reverse KL over position intervals on the student responses split by the final answer correctness.}
  \label{fig:teacher_unanchoring_example}
\end{figure}

Vanilla OPD samples trajectories from the student, and the teacher feedback is evaluated on states that may be far from where it would actually visit. We name it as \emph{teacher unanchoring}. Figure~\ref{fig:teacher_unanchoring_example} illustrates the average entropy and reverse KL over position intervals on the student (Qwen3-1.7B-Base) responses for 512 randomly selected DAPO-MATH-17K \citep{yu2025dapo} prompts, split by the final answer correctness. The teacher (Qwen3-30B-A3B non-thinking) queried by incorrect trajectories yields high uncertainty, even as the reverse KL decays along the position. We propose Interactive-Policy Distillation (IPD), which formulates generation as a state machine in which the student and teacher alternate as proposer and verifier. to address this challenge by dynamic adaptive teacher intervention during rollout. 

\subsubsection{Adaptive Teacher Intervention with Bidirectional Propose-and-Verify}

Let $m_t\in\{\mathsf{S},\mathsf{T}\}$ denote the driver model state before generating token at position $t$, where $\mathsf{S}$ and $\mathsf{T}$ indicate student driving and teacher driving, respectively, and we initialize every rollout with $m_1=\mathsf{S}$. Let $z_t\in\{\mathsf{S},\mathsf{T}\}$ record the source of the token actually emitted at position $t$. We use top-$K$ coverage as the takeover and handback rules, with $S_t^\pi$ and $S_t^q$ defined in Eq.~(\ref{eq:topk_sets}).

\noindent\textbf{Student driving.} When $m_t=\mathsf{S}$, the student proposes $\widetilde{x}_t^\pi\sim\pi_\theta(\cdot\mid c_t)$ and the teacher verifies whether the student-generated token is covered by the teacher's top-$K$ support, \emph{i.e.,} $\widetilde{x}_t^\pi \in S_t^q$. If covered, which indicates that the student's proposal within the teacher's acceptance region, IPD accepts the student proposal, assigns $z_t=\mathsf{S}$, and remains in student driving state. Otherwise, the proposal is discarded, and the teacher samples a replacement $\widetilde{x}_t^q\sim q(\cdot\mid c_t)$. IPD emits this teacher-sampled token with $z_t=\mathsf{T}$, and the rollout control switches to the teacher from position $t+1$.

\noindent\textbf{Teacher driving.} When $m_t=\mathsf{T}$, the teacher proposes $\widetilde{x}_t^q\sim q(\cdot\mid c_t)$ and the student verifies whether the teacher-sampled token satisfies $\widetilde{x}_t^q\in S_t^\pi$. If it is not covered by the student's top-$K$ support, the teacher retains control. Otherwise, as the teacher's proposal is back to the student's acceptance region, which indicates that the student is possible to generate is on its own, it is kept and serves as the handback trigger and the rollout control is returned to student rom position $t+1$.

The complete per-token transition can be written as
\begin{equation}
    (x_t,z_t,m_{t+1})=
    \begin{cases}
        (\widetilde{x}_t^\pi,\mathsf{S},\mathsf{S}), & m_t=\mathsf{S},\ \widetilde{x}_t^\pi\in S_t^q,\\
        (\widetilde{x}_t^q,\mathsf{T},\mathsf{T}), & m_t=\mathsf{S},\ \widetilde{x}_t^\pi\notin S_t^q,\\
        (\widetilde{x}_t^q,\mathsf{T},\mathsf{S}), & m_t=\mathsf{T},\ \widetilde{x}_t^q\in S_t^\pi,\\
        (\widetilde{x}_t^q,\mathsf{T},\mathsf{T}), & m_t=\mathsf{T},\ \widetilde{x}_t^q\notin S_t^\pi.
    \end{cases}
    \label{eq:ipd_state_transition}
\end{equation}
The state machine iterates until an EOS token is emitted or the maximum rollout length is reached. Eq.~(\ref{eq:ipd_state_transition}) is written token by token for clarity. Following \citet{leviathan2023fast} and \citet{skd}, for a faster rollout, the implementation usually consecutively proposes a chunk of $\gamma$ tokens and verify them in parallel, instead of token-by-token verification immediately after decoded. 

Note that in this paper we use the top-$K$ coverage as the takeover and handback rules. As a unified framework, IPD is also compatible with other rules, and the takeover and handback rules can also be asymmetric. For example, we can instead use a threshold on the entropy of the student or the teacher distributions to determine whether to take over or hand back control. SKD \citep{skd} can be included as a special case of IPD with a top-$K$ coverage takeover rule and a fixed-one-token handback rule, and in our experiments, SKD is also implemented under our customized fused engine for more efficient rollout.

\subsubsection{Source-Split KL Objective}

The state machine model formulated by Eq.~(\ref{eq:ipd_state_transition}) produces a response trajectory $x$ together consisting of student-generated and teacher-generated tokens with recorded source labels $z=(z_1,\ldots,z_{T})$. Denote $I_{on}(z)=\{t:z_t=\mathsf{S}\}$ and $I_{off}(z)=\{t:z_t=\mathsf{T}\}$ as the on-policy (student-generated) and off-policy (teacher-generated) position sets, respectively. Since the trajectory $x$ is a mixure of student and teacher tokens, it is natural provide the two groups of positions with different supervisions. We therefore design a source-split KL objective that applies different KL divergences to student-generated and teacher-generated positions.

Specifically, the token-level KL divergences applied to student-generated and teacher-generated positions are respectively denoted as 
\begin{align}
    \ell_t^{\mathrm{on}}(\theta)
    =D_{\mathrm{KL}}\!\left(\bar{\pi}^{S_t}_\theta(\cdot\mid c_t)\,\Vert\,\bar{q}^{S_t}(\cdot\mid c_t)\right), \quad
    \ell_t^{\mathrm{off}}(\theta)
    =D_{\mathrm{KL}}\!\left(\bar{q}^{S_t}(\cdot\mid c_t)\,\Vert\,\bar{\pi}^{S_t}_\theta(\cdot\mid c_t)\right). \label{eq:ipd_off_loss}
\end{align}
Student-generated positions therefore receive the token-level reverse KL used by on-policy distillation, while teacher-generated tokens receive the token-level forward KL used by off-policy distillation. Both losses are evaluated at the realized mixed-source prefix $c_t$, regardless of which model generated earlier tokens.
Let $\rho_\theta^{\mathrm{IPD}}(x,z\mid y)$ denote the joint distribution over emitted responses and source labels induced by the state machine in Eq.~(\ref{eq:ipd_state_transition}). The final length-normalized objective is minimizing the following source-split KL
\begin{equation}
    \mathcal{L}_{\mathrm{IPD}}(\theta)=\mathbb{E}_{\substack{y\sim\mathcal{D},\,(x,z)\sim\rho_\theta^{\mathrm{IPD}}(\cdot,\cdot\mid y)}}\!\left[\frac{1}{|x|}\left(\sum_{t\in I_{on}(z)}\ell_t^{\mathrm{on}}(\theta)+\sum_{t\in I_{off}(z)}\ell_t^{\mathrm{off}}(\theta)\right)\right].
    \label{eq:ipd_objective}
\end{equation}

With the bidirectional propose-and-verify mechanism and the source-split KL, IPD naturally becomes a unified bridge between on-policy and off-policy distillations. Specifically, never transferring control to the teacher yields length-normalized on-policy distillation with reverse KL supervision on pure student-generated trajectories, while always rejecting the first student proposal and never handing control back yields off-policy distillation with forward KL supervision on pure teacher-generated trajectories.

\subsubsection{Fused Inference Engine for Efficient Dual-Model Scheduling}

\begin{wrapfigure}{r}{0.45\textwidth}
  \centering
  \includegraphics[width=\linewidth]{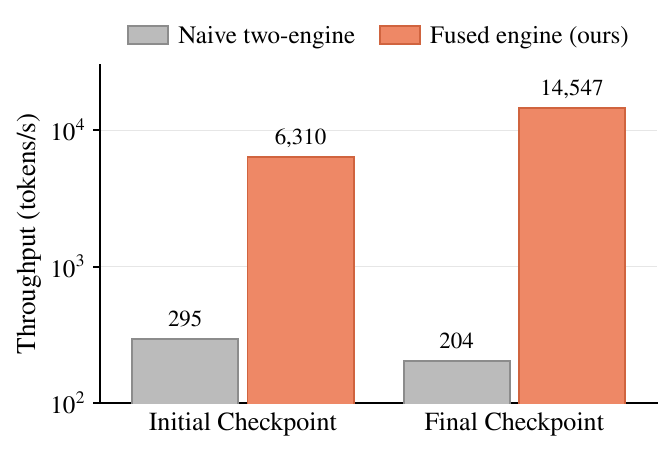}
  \setlength{\abovecaptionskip}{5pt}
  \vspace{-10pt}
  \caption{IPD rollout throughput of fused engine and naive two-engine implementation.}
  \label{fig:fused_vs_naive_throughput}
\end{wrapfigure}

To integrate IPD with verl \citep{sheng2025hybridflow}, built on vLLM \citep{kwon2023efficient}, we design a fused inference engine for more efficient dual-model scheduling. 
A naive implementation that hosts two models on separate vLLM serving instances will have to re-encode the whole prefix for every verification round, as an off-the-shelf vLLM engine bypasses its prefix cache for requests that return prompt log-probabilities and cannot append externally generated tokens to the KV cache of a live request. Besides, there are also communication overheads between two engines. 
As shown in Figure~\ref{fig:fused_vs_naive_throughput}, with Qwen3-4B-Base (non-thinking) as the teacher and Qwen3-1.7B-Base as the student, executing the IPD rollout with bidirectional propose-and-verify on the same 512 randomly selected DAPO-MATH-17K prompts under identical settings and memory budget, fused engine is 21.4$\times$ faster than the naive implementation on the initial checkpoint and 71.4$\times$ faster on the final one. 

\begin{figure}[!h]
  \includegraphics[width=0.99\textwidth]{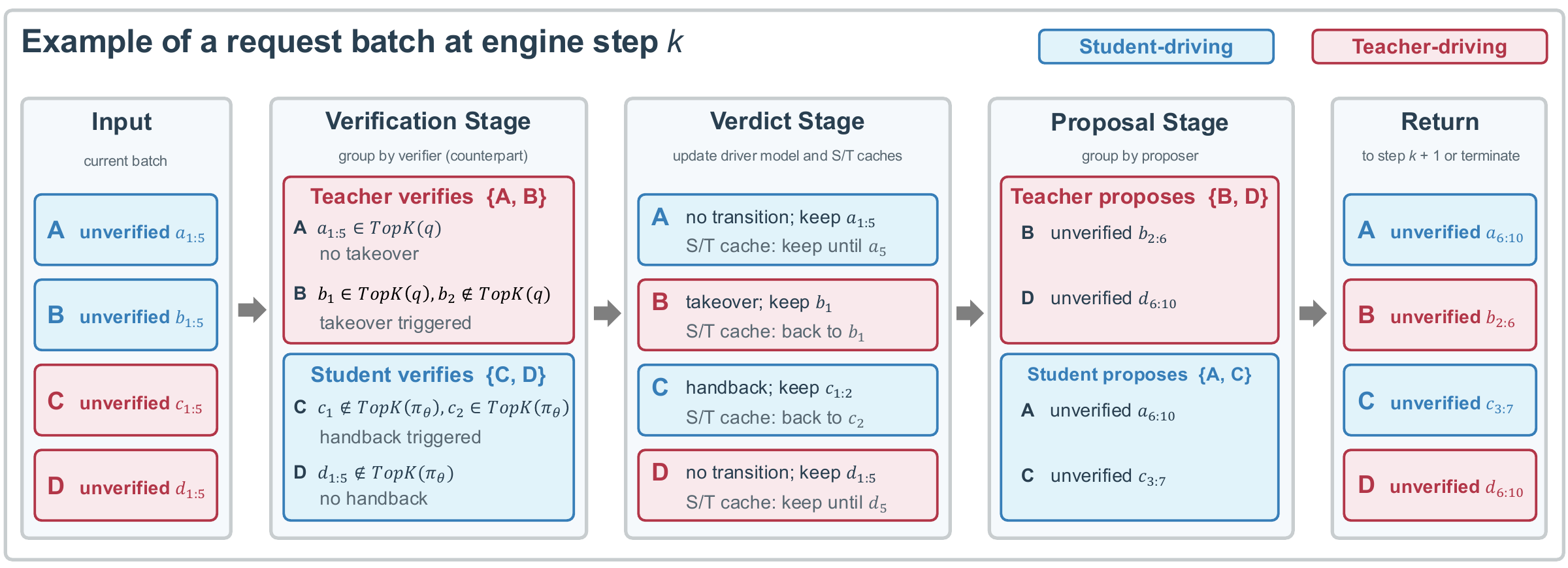}
  \caption{Example of one IPD fused engine step for a request batch. The engine distributes requests to their respective verifier models for verification, collects verification results, updates their driver models, manages their KV cache on each model, re-distributes requests to their respective proposer models for new unverified draft tokens, and return the updated requests for next step or termination.}
  \label{fig:fused_engine_step}
\end{figure}

Specifically, as illustrated in Figure~\ref{fig:fused_engine_step}, our fused engine co-hosts the student and teacher in one serving instance, and each active request maintains a verified prefix, a chunk of unverified draft tokens, the identity of its current driver model, and separate student and teacher KV-cache states. 
For each request, the two models serve as symmetrical roles. Each model maintains its own cache and may act as either the proposer or the verifier, with the verifier always being the counterpart of the current driver model.
One fused engine step consists of a verification, a verdict, and a proposal stage. In the verification stage, requests are grouped by their verifier, and each verifier evaluates all unverified draft tokens of its sub-batch in parallel through a single prefill pass. Only the newly appended tokens require computation as long as the cache of the verified prefix remains resident. If vLLM preempts a request under resource contention and evicts its cached blocks, the affected model, whether verifier or proposer, will instead recompute the KV states of the whole prefix. 
In the verdict stage, the engine determines the acceptance or rejection of each draft, detects takeover and handback transitions, updates the driver model, and aligns both KV caches with the resulting verified prefix. If no transition occurs, all drafted tokens are kept and both caches are preserved. If a student token triggers takeover, the token is rejected and will be regenerated by the teacher, so both caches are rolled back to the preceding token. If a teacher token triggers handback, the token is retained, so both caches are rolled back only to it and the subsequent draft tokens are discarded. Either way, the two caches end at the same accepted-prefix boundary. 
In the proposal stage, requests are regrouped by their updated driver model, and each proposer autoregressively decodes a new chunk of $\gamma$ draft tokens for its sub-batch while extending only its own KV cache, leaving the counterpart model to catch up through the prefill of the next verification stage. Consequently, apart from draft tokens discarded by a transition and cache blocks evicted by preemption, every token is decoded once by its proposer and prefilled once by its verifier. The engine repeats these stages until every request has emitted an accepted EOS token or reached the maximum generation length. A student-generated EOS token terminates the request only if it passes the teacher's takeover test and otherwise triggers a teacher takeover at that position, whereas a teacher-generated EOS token is accepted directly.

\section{Experiments}
\label{sec:exp}

\subsection{Experimental Settings}
We train four Qwen3 \citep{yang2025qwen3} teacher-student pairs on DAPO-Math-17K~\citep{yu2025dapo} for one epoch, with thinking mode disabled for all teachers. We set $K=16$: all methods apply distillation over the union support $S_t=S_t^\pi\cup S_t^q$, and IPD and SKD use top-$K$ coverage for takeover, with IPD also using it for handback. We use the codebase from \citet{li2026rethinking} for OPD, and additionally build our fused inference engine and implement SKD and IPD training pipelines on top of that. Both IPD and SKD use a proposal chunk length of $\gamma=5$. We evaluate the final checkpoints on six mathematical reasoning benchmarks and report mean@8 and best@8 accuracy. Appendix~\ref{sec:full_exp_details} provides more detailed model, training, hardware, and evaluation configurations.

\begin{table}[!h]
\setlength{\abovecaptionskip}{0pt}
\setlength{\aboverulesep}{0pt}
\setlength{\belowrulesep}{0pt}
\setlength{\extrarowheight}{2pt}
\centering
\small
\caption{Comparison of different distillation methods across various student and teacher model combinations. We report mean@8 and best@8 accuracy (\%) on different math reasoning benchmarks, as well as the benchmark-wise average.}
\label{tab:main_results}
\resizebox{\textwidth}{!}{%
\begin{tabular}{ll cc cc cc cc cc cc cc}
\toprule
\multicolumn{1}{c}{\multirow{2}{*}{\makecell{Teacher/\\Student}}} & \multicolumn{1}{c}{\multirow{2}{*}{Method}} & \multicolumn{2}{c}{MATH-500} & \multicolumn{2}{c}{Minerva} & \multicolumn{2}{c}{Olympiad} & \multicolumn{2}{c}{AMC23} & \multicolumn{2}{c}{AIME24} & \multicolumn{2}{c}{AIME25} & \multicolumn{2}{c}{\textbf{Avg.}} \\
\cmidrule(lr){3-4} \cmidrule(lr){5-6} \cmidrule(lr){7-8} \cmidrule(lr){9-10} \cmidrule(lr){11-12} \cmidrule(lr){13-14} \cmidrule(lr){15-16}
& & mean@8 & best@8 & mean@8 & best@8 & mean@8 & best@8 & mean@8 & best@8 & mean@8 & best@8 & mean@8 & best@8 & mean@8 & best@8 \\
\midrule
\multirow{3}{*}{\makecell[l]{Qwen3-4B/\\Qwen3-1.7B-Base}} & OPD & 66.68 & 84.06 & 26.42 & 44.04 & 31.00 & 48.45 & 35.69 & \textbf{57.29} & 7.92 & \textbf{21.77} & 2.92 & 8.70 & 28.44 & 44.05 \\
 & SKD & 66.70 & 84.50 & 27.94 & 43.69 & 29.75 & 49.52 & 35.09 & 52.90 & 8.75 & 17.41 & 4.58 & 10.73 & 28.80 & 43.13 \\
& \ipdcell{\textbf{IPD (ours)}} & \ipdcell{\textbf{69.28}} & \ipdcell{\textbf{84.59}} & \ipdcell{\textbf{29.60}} & \ipdcell{\textbf{46.26}} & \ipdcell{\textbf{31.80}} & \ipdcell{\textbf{49.63}} & \ipdcell{\textbf{36.14}} & \ipdcell{57.25} & \ipdcell{\textbf{9.17}} & \ipdcell{21.69} & \ipdcell{\textbf{6.67}} & \ipdcell{\textbf{15.06}} & \ipdcell{\textbf{30.44}} & \ipdcell{\textbf{45.75}} \\
\midrule
\multirow{3}{*}{\makecell[l]{Qwen3-30B-A3B/\\Qwen3-1.7B-Base}} & OPD & 63.35 & 83.54 & 23.25 & 41.43 & 28.14 & 46.88 & 32.68 & 54.39 & 10.42 & 18.84 & \textbf{5.83} & 13.12 & 27.28 & 43.03 \\
 & SKD & 67.38 & 84.34 & 26.70 & 44.69 & 30.39 & 49.73 & 33.89 & 54.67 & 8.75 & 22.36 & \textbf{5.83} & \textbf{14.92} & 28.82 & 45.12 \\
& \ipdcell{\textbf{IPD (ours)}} & \ipdcell{\textbf{69.35}} & \ipdcell{\textbf{84.61}} & \ipdcell{\textbf{29.60}} & \ipdcell{\textbf{46.44}} & \ipdcell{\textbf{32.56}} & \ipdcell{\textbf{51.29}} & \ipdcell{\textbf{34.79}} & \ipdcell{\textbf{59.60}} & \ipdcell{\textbf{11.67}} & \ipdcell{\textbf{22.79}} & \ipdcell{5.42} & \ipdcell{13.10} & \ipdcell{\textbf{30.56}} & \ipdcell{\textbf{46.31}} \\
\midrule
\multirow{3}{*}{\makecell[l]{Qwen3-8B/\\Qwen3-4B-Base}} & OPD & 79.63 & 90.83 & 36.53 & 52.22 & 44.80 & 60.49 & 44.28 & 66.31 & 15.83 & 25.40 & 13.75 & 29.09 & 39.14 & 54.06 \\
 & SKD & 80.43 & 91.49 & \textbf{37.91} & 51.15 & 44.71 & 61.32 & \textbf{51.51} & \textbf{72.01} & 15.42 & 24.76 & 15.00 & \textbf{29.68} & 40.83 & 55.07 \\
& \ipdcell{\textbf{IPD (ours)}} & \ipdcell{\textbf{81.45}} & \ipdcell{\textbf{92.21}} & \ipdcell{37.87} & \ipdcell{\textbf{52.66}} & \ipdcell{\textbf{46.43}} & \ipdcell{\textbf{62.19}} & \ipdcell{48.34} & \ipdcell{67.83} & \ipdcell{\textbf{17.08}} & \ipdcell{\textbf{28.86}} & \ipdcell{\textbf{17.08}} & \ipdcell{28.66} & \ipdcell{\textbf{41.38}} & \ipdcell{\textbf{55.40}} \\
\midrule
\multirow{3}{*}{\makecell[l]{Qwen3-30B-A3B/\\Qwen3-4B-Base}} & OPD & 81.38 & 90.00 & 39.11 & 53.41 & 46.66 & 61.18 & 48.80 & 67.88 & 15.42 & 24.78 & 12.92 & 26.36 & 40.71 & 53.93 \\
 & SKD & 81.03 & 91.96 & 40.40 & 55.39 & 47.42 & \textbf{63.95} & \textbf{49.85} & \textbf{72.82} & 16.25 & 24.85 & \textbf{16.67} & \textbf{27.60} & 41.93 & 56.09 \\
& \ipdcell{\textbf{IPD (ours)}} & \ipdcell{\textbf{82.28}} & \ipdcell{\textbf{92.40}} & \ipdcell{\textbf{41.18}} & \ipdcell{\textbf{55.52}} & \ipdcell{\textbf{47.66}} & \ipdcell{63.45} & \ipdcell{49.55} & \ipdcell{70.64} & \ipdcell{\textbf{19.58}} & \ipdcell{\textbf{28.39}} & \ipdcell{13.75} & \ipdcell{26.22} & \ipdcell{\textbf{42.33}} & \ipdcell{\textbf{56.10}} \\
\bottomrule
\end{tabular}}
\end{table}

\subsection{Main Results}

We compare IPD with OPD \citep{lu2025onpolicydistillation,li2026rethinking} and SKD \citep{skd} on multiple student and teacher model combinations across different sizes. Thinking mode is disabled for all non-base teachers. In Table~\ref{tab:main_results}, our algorithm demonstrates consistent improvements over the baselines across all benchmarks and model combinations. Specifically, when distilling Qwen3-4B to Qwen3-1.7B-Base, IPD achieves an average mean@8 accuracy of $30.44$ across all benchmarks, outperforming OPD and SKD by $+2.00$ and $+1.64$, respectively. Besides, IPD also allows the student to robustly learn from a larger teacher. According to the discussion in \citet{li2026rethinking}, a larger teacher cannot always distill a better student, and when the teacher-student reasoning gap is too large, the distillation performance can be suboptimal. In our experiments, when distilling Qwen3-30B-A3B to Qwen3-1.7B-Base, we also find this phenomenon that the distilled student achieves an average mean@8 accuracy of $27.28$, which is lower than the $28.44$ achieved by distilling a Qwen3-4B teacher. Meanwhile, with the bidirectional propose-and-verify mechanism, IPD can effectively bridge the reasoning gap between the teacher and student, achieving an average mean@8 accuracy of $30.56$, which is $3.28$ higher than OPD and $1.74$ higher than SKD. 

\begin{figure}[!h]
    \centering
    \includegraphics[width=0.99\linewidth]{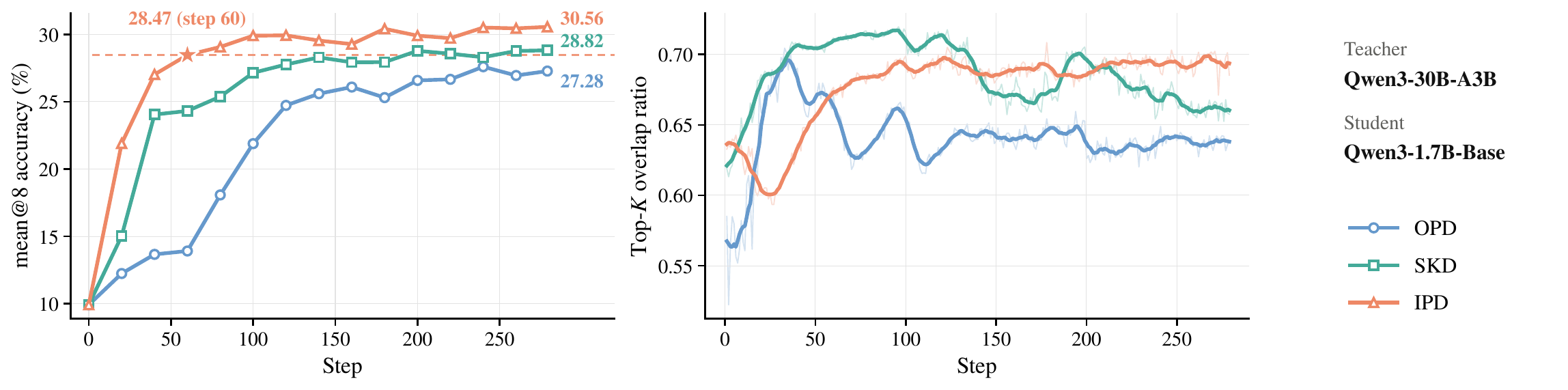}
    \caption{The mean@8 accuracy over math reasoning benchmarks and top-$K$ overlap ratio during training on DAPO-MATH-17K. In addition to better final checkpoint performance, IPD also shows a much higher data efficiency than other baselines.}
    \label{fig:main_training_curve}
\end{figure}

Besides, IPD is also a more data-efficient distillation algorithm. As shown in Figure~\ref{fig:main_training_curve}, IPD shows a much faster convergence than other baselines. Specifically, trained with only fewer than 1/4 training examples for 60 training steps, IPD can outperform OPD and matches SKD trained for a whole epoch with 279 training steps. Besides, IPD shows a stably increasing top-$K$ overlap ratio (defined by $\vert S^\pi_t \cap S^q_t \vert / K$), indicating a healthy training under large a teacher-student gap \citep{li2026rethinking}.

\subsection{Ablation Study}

\begin{figure}[!h]
  \centering
  \includegraphics[width=0.99\linewidth]{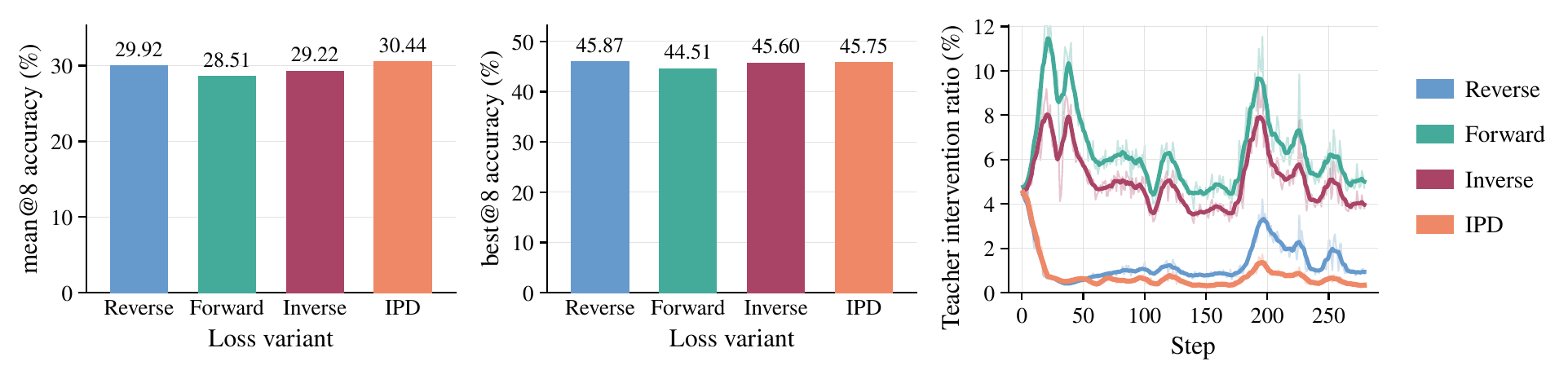}
  \caption{The mean@8 and best@8 average accuracy over benchmarks and the teacher intervention ratio with different losses during IPD training.}
  \label{fig:ablation_loss}
\end{figure}

\noindent\textbf{KL direction.} To investigate the impact of different KL directions for on-policy and off-policy positions, and the resonability of our source-split KL design, we conduct an ablation study on the IPD KL objective. We compare the following configurations: 
(i) Forward KL for all positions (Pure Forward KL), (ii) Reverse KL for all positions (Pure Reverse KL), (iii) Forward KL for on-policy and Reverse KL for off-policy positions (Inverse-IPD-Split KL), and (iv) Reverse KL for on-policy and Forward KL for off-policy positions (IPD-Split KL, default).
As shown in Figure~\ref{fig:ablation_loss}, we observe that the IPD Default configuration, which uses Reverse KL for on-policy positions and Forward KL for off-policy positions, achieves the best mean@8 accuracy. Meanwhile, Pure Reverse KL turns out to be the runner-up, with the second-best mean@8 accuracy and a $0.12$ higher best@8 accuracy than the default source-split KL. The teacher intervention ratio, which is calculated by the proportion of teacher-generated tokens among all tokens, gradually decreases during training, indicating that the student is learning to generate more tokens that are covered by the teacher's top-$K$ support, and leaning more towards on-policy paradigm.

\begin{figure}[!h]
  \centering
  \includegraphics[width=0.99\linewidth]{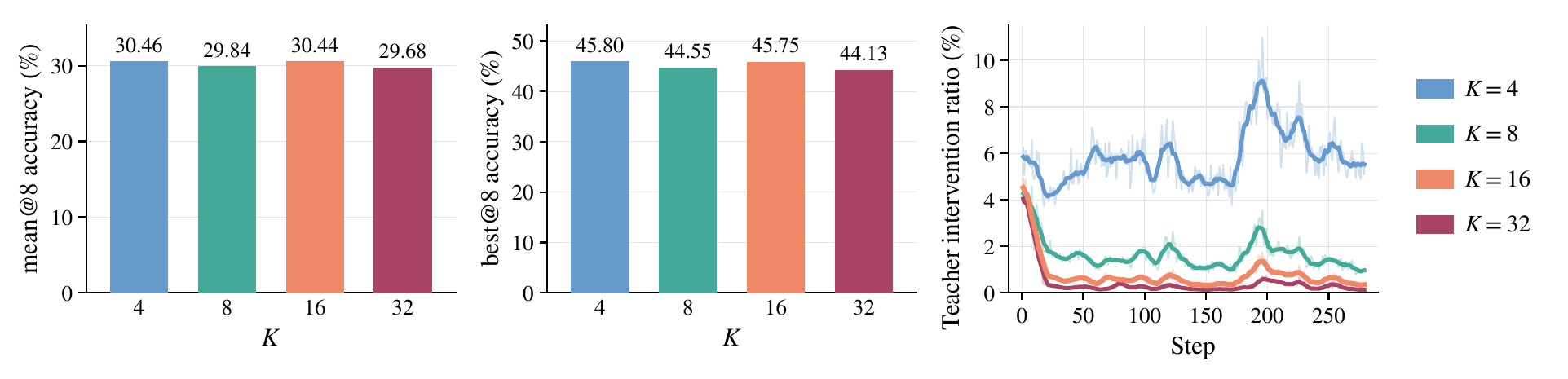}
  \caption{The mean@8 and best@8 average accuracy over benchmarks and the teacher intervention ratio with different coverage widths $K$ during IPD training. A smaller $K$ leads to more frequent teacher takeovers and less frequent teacher handbacks, resulting in a higher intervention ratio.}
  \vspace{-8pt}
  \label{fig:ablation_k_values}
\end{figure}

\noindent\textbf{Coverage width.}
In our main experiments, we set $K=16$ for the top-$K$ coverage takeover and handback rules in the IPD framework. Under a smaller $K$, there will be more teacher takeovers and less frequent teacher handbacks, which leans more towards off-policy training, results in a higher teacher invervention ratio during the rollout stage, and requires more computational resources. While under a larger $K$, the teacher will invervene less, leaning more towards an on-policy training. To study the effect of different coverage widths, we further conduct experiments with $K=4$, $K=8$, and $K=32$. 
As shown in Figure~\ref{fig:ablation_k_values}, IPD under $K=4$ achieves the best performance with $30.46$ mean@8 and $45.80$ best@8. The default $K=16$ appears to be a close runner-up with only $-0.02$ mean@8 and $-0.05$ best@8 gap, while getting much lower teacher intervention ratio, which provides a good balance between performance and computational efficiency.

\begin{figure}[!h]
  \includegraphics[width=0.99\linewidth]{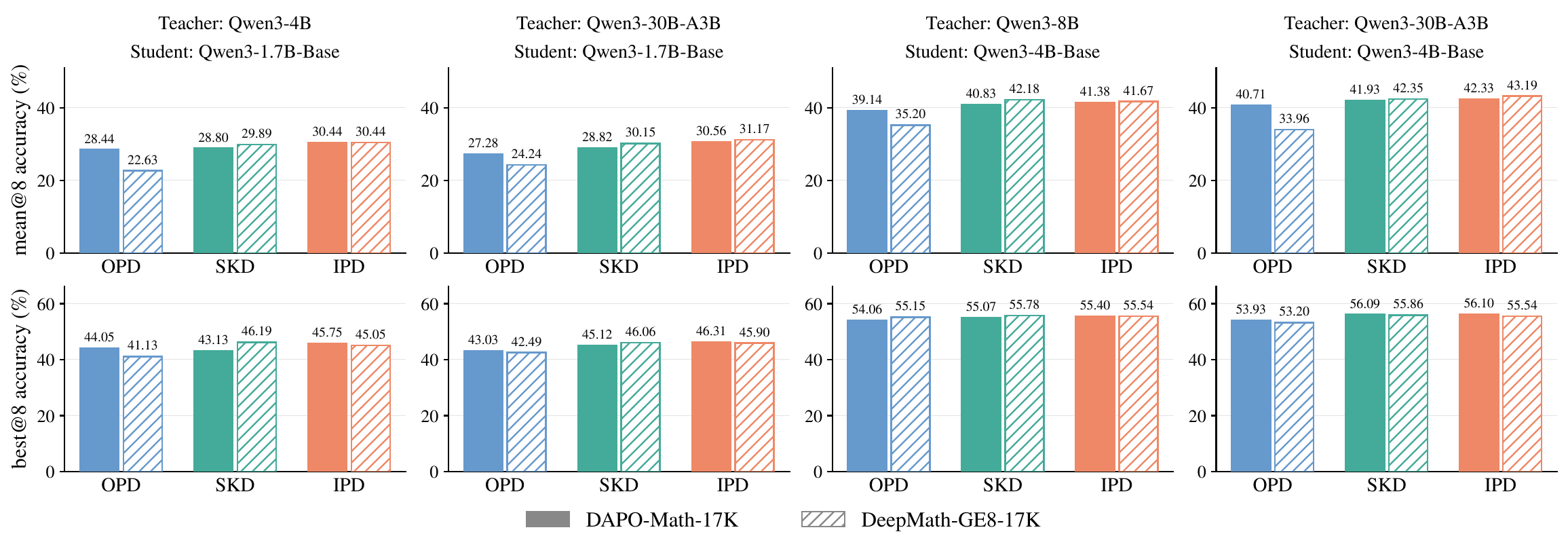}
  \caption{Comparison of different distillation algorithms on DAPO-MATH-17K and DeepMath-GE8-17K. Intervention-based distillation algorithms including IPD and SKD demonstrate consistent robustness, while OPD apprears to be more sensitive to the change of training data.}
  \label{fig:robustness_on_different_training_data}
\end{figure}

\subsection{Robustness on Different Training Data}
We further explore the robustness of IPD and other distillation algorithms under different training data. In addition to DAPO-MATH-17K, we select examples from DeepMath-103K \citep{he2026deepmath} with difficulty levels greater than or equal to 8 and get 17,085 training examples, denoted as DeepMath-GE8-17K. As shown in Figure~\ref{fig:robustness_on_different_training_data}, we compare different algorithms on these two training datasets. Specifically, we find that algorithms with teacher intervention during rollout, including IPD and SKD, demonstrate a good resistance to the change of training data. 
Meanwhile, OPD shows a non-negligible mean@8 accuracy drop, which indicates that OPD is more sensitive to the training data. On the Qwen3-30B-A3B to Qwen3-4B-Base group, the mean@8 accuracy of OPD even gets a significant drop of $-6.75$. This indicates that without teacher intervention, OPD is more sensitive to the training data, and teacher intervention applied by IPD and SKD can alleviate this sensitivity, which requires less effort on data preparation and makes them more practical paradigms.

\section{Conclusion and Discussion}
In this paper, we propose Interactive-Policy Distillation (IPD), which addresses the teacher unanchoring issue and improve the distillation performance through bidirectional takeover and handback during rollout, together with source-split supervision based on token provenance. This design dynamically corrects trajectories that leave the teacher's acceptance region, and also provides a unified bridge between on-policy and off-policy distillations. Across diverse teacher-student pairs, IPD demonstrates better distillation performance and higher data efficiency, as well as better robustness to training data. We also design a fused inference engine to support our algorithm with efficient dual-model scheduling during the rollout stage.

Currently, like most works in the OPD field, we only consider same-tokenization distillation here. Extending our method to cross-tokenization scenarios is an important problem. Besides, more different rules beyond top-$K$ coverage can be further explored. On the infrastructure side, we only consider synchrnous scheduling for verification and proposal stages. Optimizations like asynchronous scheduling can further improve the throughput of dual-model rollout. We leave these directions for future exploration.








\bibliography{iclr2027_conference}
\bibliographystyle{iclr2027_conference}

\newpage
\appendix

\section{Detailed Experimental Settings}
\label{sec:full_exp_details}

\paragraph{Data and models.} We conduct our main experiments on DAPO-Math-17K~\citep{yu2025dapo} with four teacher--student pairs: Qwen3-4B / Qwen3-1.7B-Base, Qwen3-30B-A3B / Qwen3-1.7B-Base, Qwen3-8B / Qwen3-4B-Base, and Qwen3-30B-A3B / Qwen3-4B-Base. Thinking mode is disabled for all teacher models.

\paragraph{Training configuration.} We use the codebase from \citet{li2026rethinking} for OPD, which is under the verl \citep{sheng2025hybridflow} framework. On top of their codebase, we build our fused inference engine and implement SKD and IPD training pipelines. Specifically, as a special case of IPD, SKD also uses our fused inference engine for rollout. Following \citet{li2026rethinking}, unless otherwise specified, we train each model for one epoch (279 steps) with a learning rate of $1\times10^{-6}$. The maximum rollout sequence length is 8192, comprising a maximum prompt length of 1024 and a maximum generation length of 7168. Each prompt produces four training rollouts sampled with temperature 1.0 and top-$p=1.0$. All experiments are conducted on two compute nodes, each equipped with 4 NVIDIA H100 SXM 80GB GPUs. 

\paragraph{Distillation configuration.} All methods use top-$K$ distillation with $K=16$. Each KL term is computed after separately renormalizing the student and teacher distributions over the union support $S_t=S_t^\pi\cup S_t^q$, as defined in Eq.~(\ref{eq:topk_normalization}). IPD and SKD both use $K=16$ for the takeover coverage rule and a proposal length of $\gamma=5$. IPD additionally applies the symmetric top-$K$ coverage rule for handback.

\paragraph{Evaluation.} We evaluate the final checkpoint of every run on MATH-500, Minerva, Olympiad-Bench, AMC23, AIME24, and AIME25. For each problem, we independently sample eight responses with temperature 0.7 and top-$p=0.95$. Same as training, we use a maximum generation length of 7168. We report mean@8 and best@8 accuracy for each benchmark and their benchmark-wise averages. 

\section{Further Discussions}
\subsection{Sampled-Token OPD}
\label{app:sampled_token}

Since the sampled token follows $x_t\sim\pi_\theta(\cdot\mid c_t)$, taking the expectation of Eq.~(\ref{eq:sampled_token_opd}) over $x_t$ recovers the full-vocabulary reverse KL in Eq.~(\ref{eq:full_vocab_opd}),
\begin{equation}
    \mathbb{E}_{x_t\sim\pi_\theta(\cdot\mid c_t)}\bigl[\widehat{\ell}^{\mathrm{sample}}_t\bigr]
    =\sum_{v\in\mathcal{V}}\pi_\theta(v\mid c_t)\log\frac{\pi_\theta(v\mid c_t)}{q(v\mid c_t)}
    =\ell^{\mathrm{full}}_t .
    \label{eq:sampled_token_unbiased}
\end{equation}
According to \citet{schulman2020approximating}, $\widehat{\ell}^{\mathrm{sample}}_t$ is therefore the $k_1$ estimator of the reverse KL, which is unbiased even though its individual realizations can be negative while $\ell^{\mathrm{full}}_t\geq 0$ is guaranteed.
Given the identity $\nabla_\theta\pi_\theta(v\mid c_t)=\pi_\theta(v\mid c_t)\nabla_\theta\log\pi_\theta(v\mid c_t)$,
the product rule applied to Eq.~(\ref{eq:full_vocab_opd}) gives
\begin{equation}
\begin{aligned}
    \nabla_\theta\ell^{\mathrm{full}}_t
    &=\sum_{v\in\mathcal{V}}\nabla_\theta\pi_\theta(v\mid c_t)\,
      \log\frac{\pi_\theta(v\mid c_t)}{q(v\mid c_t)}
      +\sum_{v\in\mathcal{V}}\pi_\theta(v\mid c_t)\nabla_\theta\log\frac{\pi_\theta(v\mid c_t)}{q(v\mid c_t)}\\
    &=\sum_{v\in\mathcal{V}}\nabla_\theta\pi_\theta(v\mid c_t)\,
      \log\frac{\pi_\theta(v\mid c_t)}{q(v\mid c_t)}
      +\sum_{v\in\mathcal{V}}\nabla_\theta\pi_\theta(v\mid c_t)\\
    &=\sum_{v\in\mathcal{V}}\nabla_\theta\pi_\theta(v\mid c_t)\,
      \log\frac{\pi_\theta(v\mid c_t)}{q(v\mid c_t)}
      +\nabla_\theta\sum_{v\in\mathcal{V}}\pi_\theta(v\mid c_t)\\
    &=\sum_{v\in\mathcal{V}}\pi_\theta(v\mid c_t)\log\frac{\pi_\theta(v\mid c_t)}{q(v\mid c_t)}\nabla_\theta\log\pi_\theta(v\mid c_t)
      +\nabla_\theta 1\\
    &=\mathbb{E}_{x_t\sim\pi_\theta(\cdot\mid c_t)}\Bigl[\widehat{\ell}^{\mathrm{sample}}_t\,
      \nabla_\theta\log\pi_\theta(x_t\mid c_t)\Bigr].
\end{aligned}
\label{eq:full_vocab_grad}
\end{equation}
We denote the gradient estimator as
\begin{equation}
    \widehat{g}_t=\widehat{\ell}^{\mathrm{sample}}_t\,
    \nabla_\theta\log\pi_\theta(x_t\mid c_t). 
    \label{eq:sampled_token_grad}
\end{equation}
Under the context of gradient ascent in reinforcement learning, $-\widehat{g}_t$ is the policy gradient obtained when OPD is implemented as maximizing
\begin{equation}
\begin{aligned}
  \mathcal{J}_t = \mathbb{E}_{x_t\sim\pi_\theta(\cdot\mid c_t)}\Bigl[\text{sg}\big[-\widehat{\ell}^{\mathrm{sample}}_t\big]\Bigr],
\end{aligned} 
\end{equation}
where the per-token reward is the stop-gradient operated $-\widehat{\ell}^{\mathrm{sample}}_t = \log q(x_t\mid c_t)-\log\pi_\theta(x_t\mid c_t)$ \citep{lu2025onpolicydistillation, oh2026klforkl}. 

Given a prefix $c_t$, variances of the sampled-token loss and gradient are
\begin{align}
    \operatorname{Var}\bigl[\widehat{\ell}^{\mathrm{sample}}_t\mid c_t\bigr]
    &=\sum_{v\in\mathcal{V}}\pi_\theta(v\mid c_t)
      \left(\log\frac{\pi_\theta(v\mid c_t)}{q(v\mid c_t)}\right)^2
      -\bigl(\ell^{\mathrm{full}}_t\bigr)^2
    \nonumber\\
    &=\sum_{v\in\mathcal{V}}\pi_\theta(v\mid c_t)
      \left(\log\frac{\pi_\theta(v\mid c_t)}{q(v\mid c_t)}
      -\ell^{\mathrm{full}}_t\right)^2,
    \label{eq:sampled_token_variance}\\
    \operatorname{Var}\bigl[\widehat{g}_t\mid c_t\bigr]
    &=\mathbb{E}\bigl[\|\widehat{g}_t-\nabla_\theta\ell^{\mathrm{full}}_t\|^2\mid c_t\bigr]\nonumber\\
    &=\mathbb{E}\bigl[(\widehat{\ell}^{\mathrm{sample}}_t)^2
      \|\nabla_\theta\log\pi_\theta(x_t\mid c_t)\|^2\mid c_t\bigr]
      -\|\nabla_\theta\ell^{\mathrm{full}}_t\|^2
    \nonumber\\
    &=\sum_{v\in\mathcal{V}}\pi_\theta(v\mid c_t)
      \left\|\log\frac{\pi_\theta(v\mid c_t)}{q(v\mid c_t)}
      \nabla_\theta\log\pi_\theta(v\mid c_t)
      -\nabla_\theta\ell^{\mathrm{full}}_t\right\|^2.
    \label{eq:sampled_token_grad_variance}
\end{align}

Eq.~(\ref{eq:sampled_token_variance}) and Eq.~(\ref{eq:sampled_token_grad_variance}) reveal a common source of variance in sampled-token OPD.
When the teacher and student disagree strongly, the student may assign much higher probability to tokens with $q(v\mid c_t)\ll\pi_\theta(v\mid c_t)$, yielding much larger log-ratios than on tokens where the models agree \citep{fu2026revisiting}. Sampling among these tokens makes the loss and gradient fluctuate, leading to noisy and potentially unstable updates \citep{oh2026klforkl}.

\subsection{Distillation Support} 
According to the GitHub issue discussion from the official repository of \citet{li2026rethinking}, different choices of support (including $S_t^\pi$, $S_t^q$, $S_t^\pi\cup S_t^q$, and $S_t^\pi\cap S_t^q$) do not bring significant differences for same-size distillation. From our observation, for different-size distillation, using the union of the two top-$K$ sets $S_t^\pi\cup S_t^q$ as the support for top-$K$ OPD is slightly better than using others. Unless specified, we use the union of the two top-$K$ sets by default in this paper.

\subsection{Loss Aggregation}
Following GKD~\citep{gkd} and SKD~\citep{skd}, Eq.~(\ref{eq:ipd_objective}) uses a sequence-mean-token-mean aggregation, which assigns equal weight to each response and prevents longer reasoning traces from dominating the minibatch objective. Another aggregation commonly used in OPD implementations is the global token mean, which assigns equal weight to every valid token and consequently weights each rollout in proportion to its length. This aggregation choice is orthogonal to the rollout policy, the token source, and the direction of the source-split KL. From our practice, global token-mean aggregation and sequence-mean-token-mean aggregation produce comparable performances, while the latter yields slightly more stable optimization. We therefore use sequence-mean-token-mean aggregation by default in this paper. 

\subsection{Student Models}
In this paper, we use Qwen3 Base models as student for all the experiments. Non-base Qwen3 models are trained to switch between thinking mode and non-thinking mode within single model, and this switch is contorlled by special $\langle\text{think}\rangle$ and $\langle/\text{think}\rangle$ tokens. Specifically, when disabling thinking mode, the tokenizer will append an empty $\langle\text{think}\rangle\langle/\text{think}\rangle$ block to the prompt before decoding. If using non-base Qwen3 models with thinking mode disabledas student, it can be difficult to distinguish whether the student performance gain is from successfully aligning with the teacher, or it just somehow manages to bypass the template-controlled thinking mode switch and silently activates its thinking capability obtained from previous training. Therefore, if using non-base Qwen3 models with thinking mode disabled as teacher, we think using Qwen3 Base models as student is a cleaner setting for distillation experiments. Another clean setting is to use non-base Qwen3 models with thinking mode enabled as both teacher and student, but this setting would require a much longer response length and consume more computational resources. We leave this setting for future exploration.

\section{Additional Experimental Results}

\subsection{Additional Main Results}
\begin{figure}[!h]
  \includegraphics[width=0.99\linewidth]{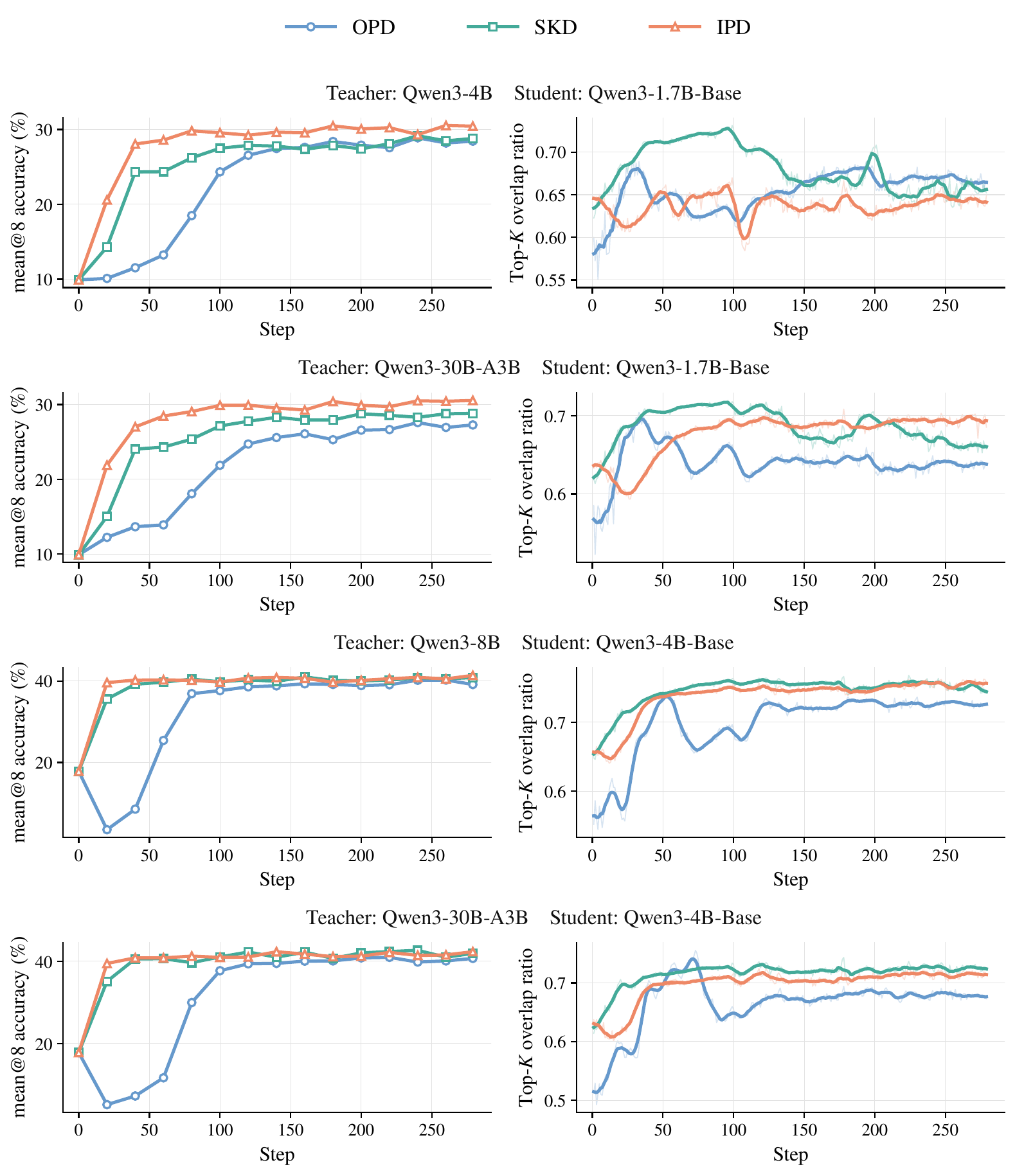}
  \caption{The mean@8 accuracy and top-$K$ overlap ratio curve for four groups of models.}
  \label{fig:qwen3_all_group_curves}
\end{figure}

In addition to the main results in Table~\ref{tab:main_results}, we also report the mean@8 accuracy and top-$K$ overlap ratio curves during training, as shown in Figure~\ref{fig:qwen3_all_group_curves}.

\FloatBarrier

\subsection{Additional Ablation Results}
\label{sec:ablation_details}

\paragraph{KL direction.} Table~\ref{tab:ablation_loss} shows that the default source-split KL obtains the highest benchmark-averaged mean@8 accuracy ($30.44$), while pure reverse KL has a slightly higher best@8 ($45.87$) than default ($45.75$). 

\begin{table}[!ht]
\centering
\small
\caption{Benchmark-wise mean@8 and best@8 accuracy for the KL-direction ablation.}
\label{tab:ablation_loss}
\resizebox{\textwidth}{!}{%
\begin{tabular}{l cc cc cc cc cc cc cc}
\toprule
& \multicolumn{2}{c}{MATH-500} & \multicolumn{2}{c}{Minerva} & \multicolumn{2}{c}{Olympiad} & \multicolumn{2}{c}{AMC23} & \multicolumn{2}{c}{AIME24} & \multicolumn{2}{c}{AIME25} & \multicolumn{2}{c}{\textbf{Avg.}} \\
\cmidrule(lr){2-3} \cmidrule(lr){4-5} \cmidrule(lr){6-7} \cmidrule(lr){8-9} \cmidrule(lr){10-11} \cmidrule(lr){12-13} \cmidrule(lr){14-15}
Loss Variant & mean@8 & best@8 & mean@8 & best@8 & mean@8 & best@8 & mean@8 & best@8 & mean@8 & best@8 & mean@8 & best@8 & mean@8 & best@8 \\
\midrule
Pure Reverse KL & 68.23 & 84.05 & 29.18 & 46.21 & \textbf{32.13} & 49.70 & 35.39 & 56.46 & \textbf{10.42} & \textbf{27.24} & 4.17 & 11.56 & 29.92 & \textbf{45.87} \\
Pure Forward KL & 66.80 & 83.64 & 28.77 & 45.15 & 30.46 & 49.69 & 34.19 & \textbf{58.58} & 6.67 & 18.41 & 4.17 & 11.60 & 28.51 & 44.51 \\
Inverse-IPD-Split KL & 67.35 & \textbf{84.84} & 27.16 & 44.06 & 31.37 & \textbf{50.31} & 33.58 & 56.63 & 9.17 & 22.66 & \textbf{6.67} & \textbf{15.09} & 29.22 & 45.60 \\
IPD-Split KL (default) & \textbf{69.28} & 84.59 & \textbf{29.60} & \textbf{46.26} & 31.80 & 49.63 & \textbf{36.14} & 57.25 & 9.17 & 21.69 & \textbf{6.67} & 15.06 & \textbf{30.44} & 45.75 \\
\bottomrule
\end{tabular}}
\end{table}

\paragraph{Coverage width.} Table~\ref{tab:ablation_k_values} shows that $K=4$ yields the highest average mean@8 and best@8, but only slightly exceeds the default $K=16$ by $+0.02$ and $+0.05$, respectively.

\begin{table}[!ht]
\centering
\small
\caption{Benchmark-wise mean@8 and best@8 accuracy for the top-$K$ coverage ablation.}
\label{tab:ablation_k_values}
\resizebox{\textwidth}{!}{%
\begin{tabular}{l cc cc cc cc cc cc cc}
\toprule
& \multicolumn{2}{c}{MATH-500} & \multicolumn{2}{c}{Minerva} & \multicolumn{2}{c}{Olympiad} & \multicolumn{2}{c}{AMC23} & \multicolumn{2}{c}{AIME24} & \multicolumn{2}{c}{AIME25} & \multicolumn{2}{c}{\textbf{Avg.}} \\
\cmidrule(lr){2-3} \cmidrule(lr){4-5} \cmidrule(lr){6-7} \cmidrule(lr){8-9} \cmidrule(lr){10-11} \cmidrule(lr){12-13} \cmidrule(lr){14-15}
Coverage width & mean@8 & best@8 & mean@8 & best@8 & mean@8 & best@8 & mean@8 & best@8 & mean@8 & best@8 & mean@8 & best@8 & mean@8 & best@8 \\
\midrule
$K=4$ & 68.30 & \textbf{85.36} & 28.49 & 43.59 & 31.97 & 49.76 & 35.24 & 53.75 & \textbf{10.00} & \textbf{24.21} & \textbf{8.75} & \textbf{18.17} & \textbf{30.46} & \textbf{45.80} \\
$K=8$ & 69.20 & 84.94 & 28.95 & 45.00 & 32.45 & \textbf{50.97} & 33.43 & 54.60 & 8.75 & 17.68 & 6.25 & 14.10 & 29.84 & 44.55 \\
\textbf{$K=16$}~~\small\textbf{(default)} & \textbf{69.28} & 84.59 & \textbf{29.60} & \textbf{46.26} & 31.80 & 49.63 & \textbf{36.14} & \textbf{57.25} & 9.17 & 21.69 & 6.67 & 15.06 & 30.44 & 45.75 \\
$K=32$ & 68.03 & 83.81 & 28.77 & 44.66 & \textbf{32.50} & 50.50 & 34.64 & 53.96 & 8.75 & 17.58 & 5.42 & 14.30 & 29.68 & 44.13 \\
\bottomrule
\end{tabular}}
\end{table}
\FloatBarrier

\section{Same-origin Distillation}
\begin{figure}[!h]
  \includegraphics[width=0.99\linewidth]{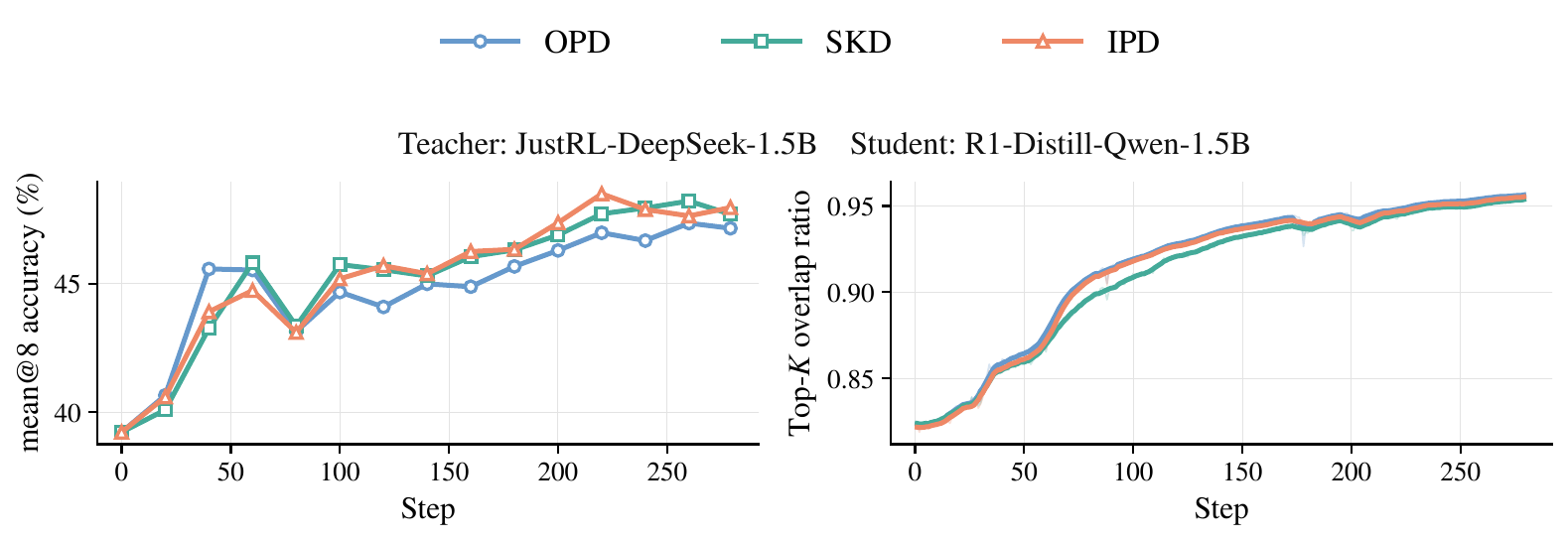}
  \caption{The mean@8 accuracy and top-$K$ overlap ratio curve for same-origin distillation experiment with DeepSeek-R1-Distill-Qwen-1.5B as the student and JustRL-1.5B as the teacher. }
  \label{fig:same_origin}
\end{figure}

In addition to different models from the Qwen3 family, we also conduct a same-origin distillation experiment with DeepSeek-R1-Distill-Qwen-1.5B as the student and JustRL-DeepSeek-1.5B as the teacher. Specifically, JustRL-DeepSeek-1.5B is the further post-trained version of DeepSeek-R1-Distill-Qwen-1.5B via simple GRPO recipe \citep{he2025justrl}. The results are shown in Table~\ref{tab:same_origin} and Figure~\ref{fig:same_origin}. We find that intervention methods including IPD and SKD do not bring as statistically significant improvement over OPD compared with our main experiments with different-size Qwen3 models. From Figure~\ref{fig:same_origin}, we can observe that the mean@8 accuracy and top-$K$ overlap ratio curves of three methods are very close to each other's, which is expected as the student and teacher are same-origin models. The same-origin experiment indicates that OPD can already achieve a good distillation performance when the reasoning gap between the student and teacher are not significant, which is consistent with the discussion from \citet{li2026rethinking}, and the main advantage of IPD remains in large-gap distillation scenarios.

\begin{table}[!h]
\setlength{\aboverulesep}{0pt}
\setlength{\belowrulesep}{0pt}
\setlength{\extrarowheight}{2pt}
\centering
\small
\caption{Same-origin distillation experiment with DeepSeek-R1-Distill-Qwen-1.5B as the student and JustRL-DeepSeek-1.5B as the teacher on DAPO-Math-17K-Processed. Intervention methods including IPD and SKD do not bring as statistically significant improvement over OPD compared with our main experiments with different-size Qwen3 models.}
\label{tab:same_origin}
\resizebox{\textwidth}{!}{%
\begin{tabular}{ll cc cc cc cc cc cc cc}
\toprule
\multicolumn{1}{c}{\multirow{2}{*}{\makecell{Teacher/\\Student}}} & \multicolumn{1}{c}{\multirow{2}{*}{Method}} & \multicolumn{2}{c}{MATH-500} & \multicolumn{2}{c}{Minerva} & \multicolumn{2}{c}{Olympiad} & \multicolumn{2}{c}{AMC23} & \multicolumn{2}{c}{AIME24} & \multicolumn{2}{c}{AIME25} & \multicolumn{2}{c}{\textbf{Avg.}} \\
\cmidrule(lr){3-4} \cmidrule(lr){5-6} \cmidrule(lr){7-8} \cmidrule(lr){9-10} \cmidrule(lr){11-12} \cmidrule(lr){13-14} \cmidrule(lr){15-16}
& & mean@8 & best@8 & mean@8 & best@8 & mean@8 & best@8 & mean@8 & best@8 & mean@8 & best@8 & mean@8 & best@8 & mean@8 & best@8 \\
\midrule
\multirow{3}{*}{\makecell[l]{JustRL-DeepSeek-1.5B/\\DeepSeek-R1-Distill-\\Qwen-1.5B}} & OPD & 85.15 & 92.79 & \textbf{38.01} & 49.41 & 47.29 & 60.63 & \textbf{64.16} & \textbf{82.92} & 27.50 & \textbf{52.85} & 20.83 & 30.59 & 47.16 & \textbf{61.53} \\
& SKD & 85.55 & \textbf{93.36} & 37.13 & 47.80 & \textbf{48.74} & \textbf{61.70} & 63.55 & 79.67 & 27.92 & 50.43 & \textbf{23.33} & \textbf{32.58} & 47.70 & 60.92 \\
& \ipdcell{\textbf{IPD (ours)}} & \ipdcell{\textbf{85.75}} & \ipdcell{93.29} & \ipdcell{37.78} & \ipdcell{\textbf{49.99}} & \ipdcell{48.11} & \ipdcell{61.36} & \ipdcell{62.80} & \ipdcell{79.53} & \ipdcell{\textbf{30.42}} & \ipdcell{47.13} & \ipdcell{22.92} & \ipdcell{29.80} & \ipdcell{\textbf{47.96}} & \ipdcell{60.18} \\
\bottomrule
\end{tabular}}
\end{table}

\section{Distillation on Limited Trainig Examples}
\begin{figure}[!h]
  \includegraphics[width=0.99\linewidth]{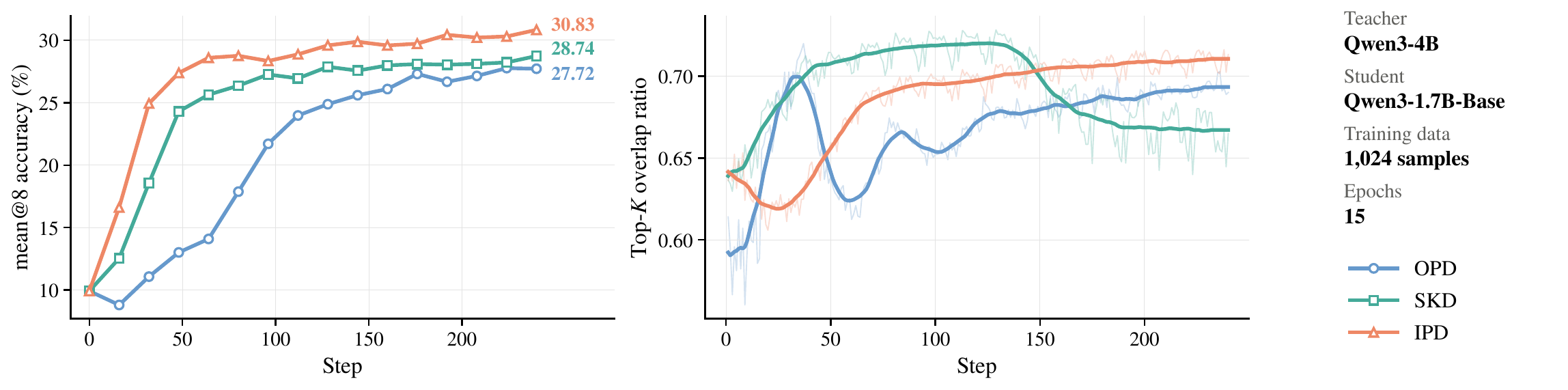}
  \caption{The mean@8 accuracy and top-$K$ overlap ratio curve for distillation on 1,024 randomly selected DAPO-MATH-17K training examples. As long as the limited unique training prompts can be replayed to generate diverse rollouts, the distillation can still be effective.}
  \label{fig:limited_data}
\end{figure}

We also run distillation trainings on limited training examples to evaluate the impact of data scarcity on the distillation process. Specifically, we randomly select 1,024 training examples from DAPO-MATH-17K and conduct distillation experiments using Qwen3-4B as teacher and Qwen3-1.7B-Base as student. The student model is trained for 15 epochs (with 240 steps in total), and the other training configurations remains the same as the main experiments. 
The results are shown in Figure~\ref{fig:limited_data}. We find that IPD can still outperform OPD and SKD and shows a higher convergence under limited training data. Another interesting observation is that, even if the training examples only count for about 1/17 of the original training data, all the three methods can matches the performance of the student model trained on the full training dataset. The limited training data experiment indicates that although with limited unique training prompts, as long as they can be replayed to generate diverse rollouts, the distillation can still be effective. 

\end{document}

%% file: math_commands.tex
\usepackage{amsmath,amsfonts,bm}

\def\eqref#1{equation~\ref{#1}}

\def\1{\bm{1}}

\DeclareMathAlphabet{\mathsfit}{\encodingdefault}{\sfdefault}{m}{sl}
\SetMathAlphabet{\mathsfit}{bold}{\encodingdefault}{\sfdefault}{bx}{n}



%% file: iclr2027_conference.bib
@article{hinton2015distilling,
  title={Distilling the knowledge in a neural network},
  author={Hinton, Geoffrey and Vinyals, Oriol and Dean, Jeff},
  journal={arXiv preprint arXiv:1503.02531},
  year={2015}
}

@inproceedings{kim2016sequence,
  title={Sequence-level knowledge distillation},
  author={Kim, Yoon and Rush, Alexander M},
  booktitle={Proceedings of the 2016 conference on empirical methods in natural language processing},
  pages={1317--1327},
  year={2016}
}

@inproceedings{
ko2025distillm,
title={Disti{LLM}-2: A Contrastive Approach Boosts the Distillation of {LLM}s},
author={Jongwoo Ko and Tianyi Chen and Sungnyun Kim and Tianyu Ding and Luming Liang and Ilya Zharkov and Se-Young Yun},
booktitle={Forty-second International Conference on Machine Learning},
year={2025}
}

@article{he2025justrl,
  title={Justrl: Scaling a 1.5 b llm with a simple rl recipe},
  author={He, Bingxiang and Qu, Zekai and Liu, Zeyuan and Chen, Yinghao and Zuo, Yuxin and Qian, Cheng and Zhang, Kaiyan and Chen, Weize and Xiao, Chaojun and Cui, Ganqu and others},
  journal={arXiv preprint arXiv:2512.16649},
  year={2025}
}

@inproceedings{mirzadeh2020improved,
  title={Improved knowledge distillation via teacher assistant},
  author={Mirzadeh, Seyed Iman and Farajtabar, Mehrdad and Li, Ang and Levine, Nir and Matsukawa, Akihiro and Ghasemzadeh, Hassan},
  booktitle={Proceedings of the AAAI conference on artificial intelligence},
  volume={34},
  number={04},
  pages={5191--5198},
  year={2020}
}

@inproceedings{shing2025taid,
title={{TAID}: Temporally Adaptive Interpolated Distillation for Efficient Knowledge Transfer in Language Models},
author={Makoto Shing and Kou Misaki and Han Bao and Sho Yokoi and Takuya Akiba},
booktitle={The Thirteenth International Conference on Learning Representations},
year={2025}
}

@article{bengio2015scheduled,
  title={Scheduled sampling for sequence prediction with recurrent neural networks},
  author={Bengio, Samy and Vinyals, Oriol and Jaitly, Navdeep and Shazeer, Noam},
  journal={Advances in neural information processing systems},
  volume={28},
  year={2015}
}

@article{ranzato2015sequence,
  title={Sequence level training with recurrent neural networks},
  author={Ranzato, Marc'Aurelio and Chopra, Sumit and Auli, Michael and Zaremba, Wojciech},
  journal={arXiv preprint arXiv:1511.06732},
  year={2015}
}

@inproceedings{arora2022exposure,
  title={Why exposure bias matters: An imitation learning perspective of error accumulation in language generation},
  author={Arora, Kushal and El Asri, Layla and Bahuleyan, Hareesh and Cheung, Jackie Chi Kit},
  booktitle={Findings of the Association for Computational Linguistics: ACL 2022},
  pages={700--710},
  year={2022}
}

@inproceedings{ho2023large,
  title={Large language models are reasoning teachers},
  author={Ho, Namgyu and Schmid, Laura and Yun, Se-Young},
  booktitle={Proceedings of the 61st annual meeting of the association for computational linguistics (volume 1: long papers)},
  pages={14852--14882},
  year={2023}
}

@inproceedings{magister2023teaching,
  title={Teaching small language models to reason},
  author={Magister, Lucie Charlotte and Mallinson, Jonathan and Adamek, Jakub and Malmi, Eric and Severyn, Aliaksei},
  booktitle={Proceedings of the 61st Annual Meeting of the Association for Computational Linguistics (Volume 2: Short Papers)},
  pages={1773--1781},
  year={2023}
}

@inproceedings{hsieh2023distilling,
  title={Distilling step-by-step! outperforming larger language models with less training data and smaller model sizes},
  author={Hsieh, Cheng-Yu and Li, Chun-Liang and Yeh, Chih-Kuan and Nakhost, Hootan and Fujii, Yasuhisa and Ratner, Alex and Krishna, Ranjay and Lee, Chen-Yu and Pfister, Tomas},
  booktitle={Findings of the association for computational linguistics: ACL 2023},
  pages={8003--8017},
  year={2023}
}

@inproceedings{ross2011reduction,
  title={A reduction of imitation learning and structured prediction to no-regret online learning},
  author={Ross, St{\'e}phane and Gordon, Geoffrey and Bagnell, Drew},
  booktitle={Proceedings of the fourteenth international conference on artificial intelligence and statistics},
  pages={627--635},
  year={2011},
  organization={JMLR Workshop and Conference Proceedings}
}

@inproceedings{kelly2019hg,
  title={Hg-dagger: Interactive imitation learning with human experts},
  author={Kelly, Michael and Sidrane, Chelsea and Driggs-Campbell, Katherine and Kochenderfer, Mykel J},
  booktitle={2019 International Conference on Robotics and Automation (ICRA)},
  pages={8077--8083},
  year={2019},
  organization={IEEE}
}

@inproceedings{menda2019ensembledagger,
  title={Ensembledagger: A bayesian approach to safe imitation learning},
  author={Menda, Kunal and Driggs-Campbell, Katherine and Kochenderfer, Mykel J},
  booktitle={2019 IEEE/RSJ International Conference on Intelligent Robots and Systems (IROS)},
  pages={5041--5048},
  year={2019},
  organization={IEEE}
}

@inproceedings{gu2024minillm,
  title={Minillm: Knowledge distillation of large language models},
  author={Gu, Yuxian and Dong, Li and Wei, Furu and Huang, Minlie},
  booktitle={International Conference on Learning Representations},
  volume={2024},
  pages={32694--32717},
  year={2024}
}

@inproceedings{ko2024distillm,
  title={DistiLLM: Towards Streamlined Distillation for Large Language Models},
  author={Ko, Jongwoo and Kim, Sungnyun and Chen, Tianyi and Yun, Se-Young},
  booktitle={International Conference on Machine Learning},
  pages={24872--24895},
  year={2024},
  organization={PMLR}
}

@inproceedings{li2025small,
  title={Small models struggle to learn from strong reasoners},
  author={Li, Yuetai and Yue, Xiang and Xu, Zhangchen and Jiang, Fengqing and Niu, Luyao and Lin, Bill Yuchen and Ramasubramanian, Bhaskar and Poovendran, Radha},
  booktitle={Findings of the Association for Computational Linguistics: ACL 2025},
  pages={25366--25394},
  year={2025}
}

@misc{openai2026astra,
  author = {{OpenAI}},
  title  = {{GPT-6 Astra}: A new generation of intelligence},
  year   = {2026},
  url    = {https://openai.com/index/gpt-6-astra/}
}

@misc{anthropic2026fable,
  author = {{Anthropic}},
  title  = {Introducing {Claude Fable 5.1} and {Claude Mythos 5.1}},
  year   = {2026},
  url    = {https://www.anthropic.com/claude-fable-and-mythos-5-1}
}

@misc{zai2026glm53,
  author = {{Z.ai}},
  title  = {{GLM-5.3: Frontier Coding with Emergent Cyber Capabilities}},
  year   = {2026},
  url    = {https://z.ai/blog/glm-5.3}
}

@article{lu2025onpolicydistillation,
  author = {Kevin Lu and Thinking Machines Lab},
  title = {On-Policy Distillation},
  journal = {Thinking Machines Lab: Connectionism},
  year = {2025},
  note = {https://thinkingmachines.ai/blog/on-policy-distillation},
  doi = {10.64434/tml.20251026},
}

@article{guo2025deepseek,
  title={Deepseek-r1: Incentivizing reasoning capability in llms via reinforcement learning},
  author={Guo, Daya and Yang, Dejian and Zhang, Haowei and Song, Junxiao and Wang, Peiyi and Zhu, Qihao and Xu, Runxin and Zhang, Ruoyu and Ma, Shirong and Bi, Xiao and others},
  journal={arXiv preprint arXiv:2501.12948},
  year={2025}
}

@article{xu2026deepseek,
  title={Deepseek-v4: Towards highly efficient million-token context intelligence},
  author={Xu, Anyi and Lin, Bangcai and Xue, Bing and Wang, Bingxuan and Xu, Bingzheng and Wu, Bochao and Zhang, Bowei and Lin, Chaofan and Dong, Chen and Ling, Chenchen and others},
  journal={arXiv preprint arXiv:2606.19348},
  year={2026}
}

@article{xiao2026mimo,
  title={Mimo-v2-flash technical report},
  author={Xiao, Bangjun and Xia, Bingquan and Yang, Bo and Gao, Bofei and Shen, Bowen and Zhang, Chen and He, Chenhong and Lou, Chiheng and Luo, Fuli and Wang, Gang and others},
  journal={arXiv preprint arXiv:2601.02780},
  year={2026}
}

@article{yang2025qwen3,
  title={Qwen3 technical report},
  author={Yang, An and Li, Anfeng and Yang, Baosong and Zhang, Beichen and Hui, Binyuan and Zheng, Bo and Yu, Bowen and Gao, Chang and Huang, Chengen and Lv, Chenxu and others},
  journal={arXiv preprint arXiv:2505.09388},
  year={2025}
}

@article{yu2025dapo,
  title={Dapo: An open-source llm reinforcement learning system at scale},
  author={Yu, Qiying and Zhang, Zheng and Zhu, Ruofei and Yuan, Yufeng and Zuo, Xiaochen and Yue, Yu and Dai, Weinan and Fan, Tiantian and Liu, Gaohong and Liu, Lingjun and others},
  journal={Advances in Neural Information Processing Systems},
  volume={38},
  pages={113222--113244},
  year={2025}
}

@inproceedings{he2026deepmath,
  title={Deepmath-103k: A large-scale, challenging, decontaminated, and verifiable mathematical dataset for advancing reasoning},
  author={He, Zhiwei and Liang, Tian and Xu, Jiahao and Liu, Qiuzhi and Chen, Xingyu and Wang, Yue and Song, Linfeng and Yu, Dian and Liang, Zhenwen and Wang, Wenxuan and others},
  booktitle={International Conference on Learning Representations},
  volume={2026},
  pages={138306--138322},
  year={2026}
}

@article{ma2026mopd,
  title={Mopd: Multi-teacher on-policy distillation for capability integration in llm post-training},
  author={Ma, Wenhan and Wei, Jianyu and Zhao, Liang and Zhang, Hailin and Xiao, Bangjun and Li, Lei and Yang, Qibin and Gao, Bofei and Wang, Yudong and Li, Rang and others},
  journal={arXiv preprint arXiv:2606.30406},
  year={2026}
}

@inproceedings{
li2026rethinking,
title={Rethinking On-Policy Distillation of Large Language Models: Phenomenology, Mechanism, and Recipe},
author={Yaxuan Li and Yuxin Zuo and Bingxiang He and Jinqian Zhang and Chaojun Xiao and Cheng Qian and Tianyu Yu and Huan-ang Gao and Wenkai Yang and Zhiyuan Liu and Ning Ding},
booktitle={ICML 2026 Workshop on Foundations of Deep Generative Models: Understanding Memorization, Generalization, and Reasoning},
year={2026}
}

@inproceedings{
oh2026klforkl,
title={{KL} for a {KL}: On-Policy Distillation with Control Variate Baseline},
author={Minjae Oh and Sangjun Song and Gyubin Choi and Yunho Choi and Yohan Jo},
booktitle={3rd AI for Math Workshop: Toward Self-Evolving Scientific Agents},
year={2026}
}

@article{xing2026trust,
  title={Trust Region On-Policy Distillation},
  author={Xing, Xingrun and Wang, Haoqing and Gao, Boyan and Li, Ziheng and Tang, Yehui},
  journal={arXiv preprint arXiv:2606.01249},
  year={2026}
}

@inproceedings{skd,
  title={Speculative knowledge distillation: Bridging the teacher-student gap through interleaved sampling},
  author={Xu, Wenda and Han, Rujun and Wang, Zifeng and Le, Long and Madeka, Dhruv and Li, Lei and Wang, William and Agarwal, Rishabh and Lee, Chen-Yu and Pfister, Tomas},
  booktitle={International Conference on Learning Representations},
  volume={2025},
  pages={64616--64646},
  year={2025}
}

@inproceedings{jin2026entropyaware,
title={Entropy-Aware On-Policy Distillation of Language Models},
author={Woogyeol Jin and Taywon Min and Yongjin Yang and Dennis Wei and Yi Zhou and Swanand Ravindra Kadhe and Nathalie Baracaldo and Kimin Lee},
booktitle={Forty-third International Conference on Machine Learning},
year={2026}
}

@inproceedings{
fu2026revisiting,
title={Revisiting On-Policy Distillation: Empirical Failure Modes and Simple Fixes},
author={Yuqian Fu and Haohuan Huang and Kaiwen Jiang and Jiacai Liu and Zhuo Jiang and Yuanheng Zhu and Dongbin Zhao},
booktitle={Third Conference on Language Modeling},
year={2026}
}

@article{armandpour2026unmasking,
  title={Unmasking on-policy distillation: Where it helps, where it hurts, and why},
  author={Armandpour, Mohammadreza and Ilhan, Fatih and Harrison, David and Jaiswal, Ajay and Hoang, Duc NM and Faghri, Fartash and Zhang, Yizhe and Cho, Minsik and Farajtabar, Mehrdad},
  journal={arXiv preprint arXiv:2605.10889},
  year={2026}
}

@inproceedings{leviathan2023fast,
  title={Fast inference from transformers via speculative decoding},
  author={Leviathan, Yaniv and Kalman, Matan and Matias, Yossi},
  booktitle={International Conference on Machine Learning},
  pages={19274--19286},
  year={2023},
  organization={PMLR}
}

@article{chen2023accelerating,
  title={Accelerating large language model decoding with speculative sampling},
  author={Chen, Charlie and Borgeaud, Sebastian and Irving, Geoffrey and Lespiau, Jean-Baptiste and Sifre, Laurent and Jumper, John},
  journal={arXiv preprint arXiv:2302.01318},
  year={2023}
}

@inproceedings{kwon2023efficient,
  title={Efficient memory management for large language model serving with pagedattention},
  author={Kwon, Woosuk and Li, Zhuohan and Zhuang, Siyuan and Sheng, Ying and Zheng, Lianmin and Yu, Cody Hao and Gonzalez, Joseph and Zhang, Hao and Stoica, Ion},
  booktitle={Proceedings of the 29th symposium on operating systems principles},
  pages={611--626},
  year={2023}
}

@inproceedings{gkd,
  title={On-policy distillation of language models: Learning from self-generated mistakes},
  author={Agarwal, Rishabh and Vieillard, Nino and Zhou, Yongchao and Stanczyk, Piotr and Ramos Garea, Sabela and Geist, Matthieu and Bachem, Olivier},
  booktitle={International Conference on Learning Representations},
  volume={2024},
  pages={21246--21263},
  year={2024}
}

@inproceedings{sheng2025hybridflow,
  title={Hybridflow: A flexible and efficient rlhf framework},
  author={Sheng, Guangming and Zhang, Chi and Ye, Zilingfeng and Wu, Xibin and Zhang, Wang and Zhang, Ru and Peng, Yanghua and Lin, Haibin and Wu, Chuan},
  booktitle={Proceedings of the Twentieth European Conference on Computer Systems},
  pages={1279--1297},
  year={2025}
}

@misc{schulman2020approximating,
  author       = {Schulman, John},
  title        = {Approximating {KL} Divergence},
  year         = {2020},
  howpublished = {\url{http://joschu.net/blog/kl-approx.html}}
}
